\pdfoutput=1

\documentclass[11pt]{article}

\usepackage{EMNLP2022}
\usepackage{enumitem}
\usepackage{listings}
\usepackage{xcolor}

\definecolor{codegreen}{rgb}{0,0.6,0}
\definecolor{codegray}{rgb}{0.5,0.5,0.5}
\definecolor{codepurple}{rgb}{0.58,0,0.82}
\definecolor{backcolour}{rgb}{0.95,0.95,0.92}

\lstdefinestyle{mystyle}{
    commentstyle=\color{codegreen},
    keywordstyle=\color{blue},
    numberstyle=\tiny\color{codegray},
    stringstyle=\color{codepurple},
    basicstyle=\ttfamily\footnotesize,
    breakatwhitespace=false,         
    breaklines=true,                 
    captionpos=b,                    
    keepspaces=true,                 
    numbers=left,                    
    numbersep=5pt,                  
    showspaces=false,                
    showstringspaces=false,
    showtabs=false,                  
    tabsize=2
}

\usepackage{times}
\usepackage{latexsym}
\usepackage[T1]{fontenc}
\usepackage[utf8]{inputenc}

\usepackage{algorithm}
\usepackage[noend]{algpseudocode}
\algrenewcommand\algorithmicuntil{\textbf{until}}
\algdef{SE}[DOWHILE]{Do}{doUntil}{\algorithmicdo}[1]{\algorithmicuntil\ #1}%

\usepackage{caption,subcaption}

\usepackage{graphicx}
\usepackage{tikz}
\usetikzlibrary{shadows, positioning, calc, arrows, matrix}
\usetikzlibrary{shapes, shapes.misc, shapes.multipart, shapes.arrows, shapes.callouts}
\usetikzlibrary{automata}
\usepackage{tikz-dependency}
\usepackage{multirow}
\usepackage{rotating}
\usepackage{bm}
\usepackage{amsmath}
\usepackage{amssymb}
\usepackage{amsthm}
\usepackage{color}
\usepackage{xcolor}

\definecolor{brightcerulean}{rgb}{0.11, 0.67, 0.84}
\definecolor{brickred}{rgb}{0.8, 0.25, 0.33}

\newtheorem{theorem}{Theorem}

\newcommand{\switchCameraReady}[2]{\ref{#1}}

\newcommand{\expectation}[1]{\mathop{\mathbb{E}}\left[#1\right]}

\usepackage{microtype}

\newcommand{\plotwidth}{0.405\textwidth}

\newcommand{\figureNormalSpeed}{
  \begin{figure*}[h!]
      \centering
      \subcaptionbox{Random weights}{
        \includegraphics[width=\plotwidth]{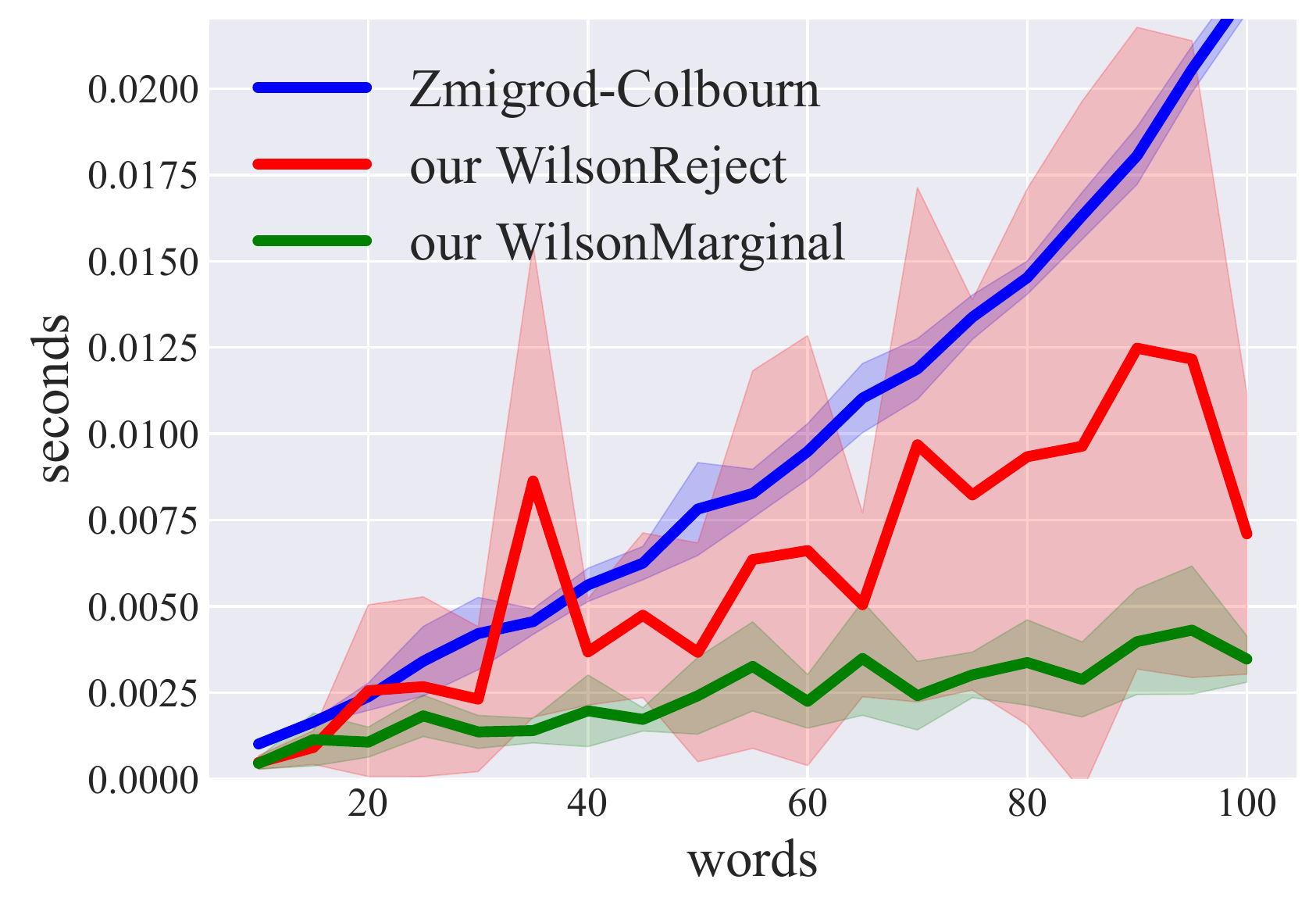}
      }
      \subcaptionbox{Trained weights}{
        \includegraphics[width=\plotwidth]{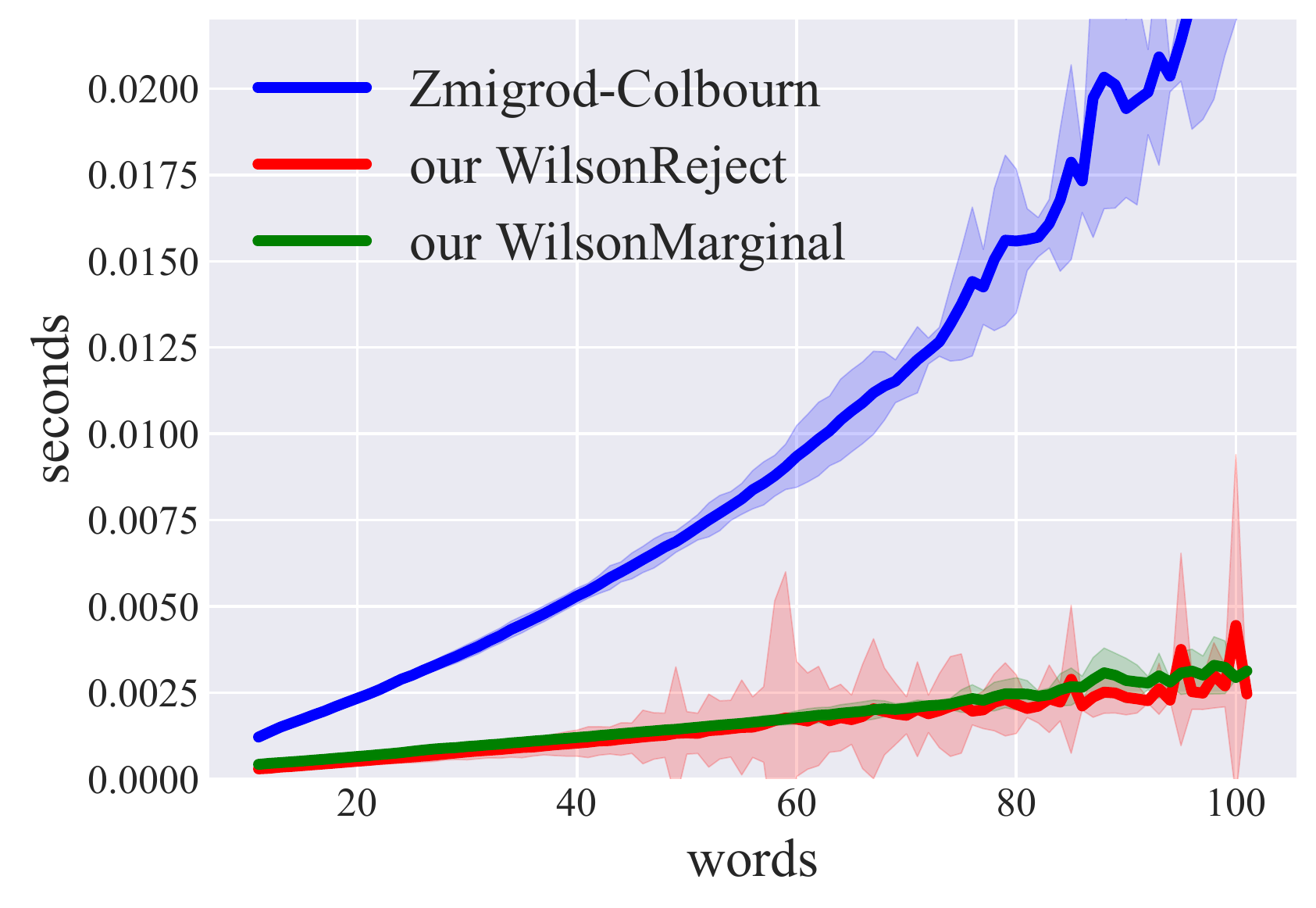}
      }
      \caption{Speed for sampling with replacement.}
      \label{fig:normal:speed}
  \end{figure*}
}

\newcommand{\figureSworSpeed}{
  \begin{figure}[t]
      \centering
      \includegraphics[width=\plotwidth]{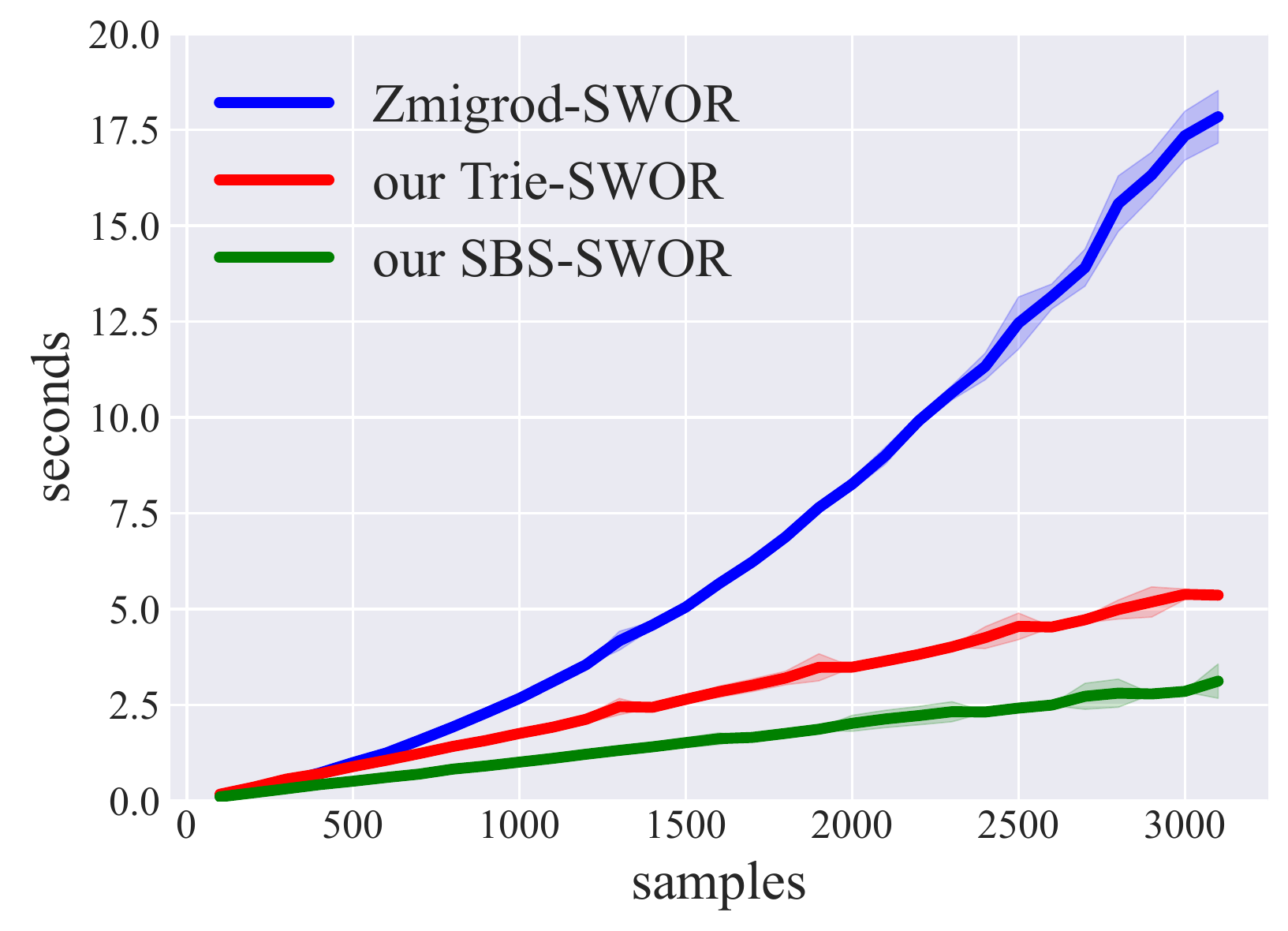}
      \caption{Speed for SWOR sampling with $14$ words.}
      \label{fig:swor:speed}
  \end{figure}
}

\newcommand{\figureGPUSpeedSimple}{
  \begin{figure}[t]
      \centering
      \includegraphics[width=\plotwidth]{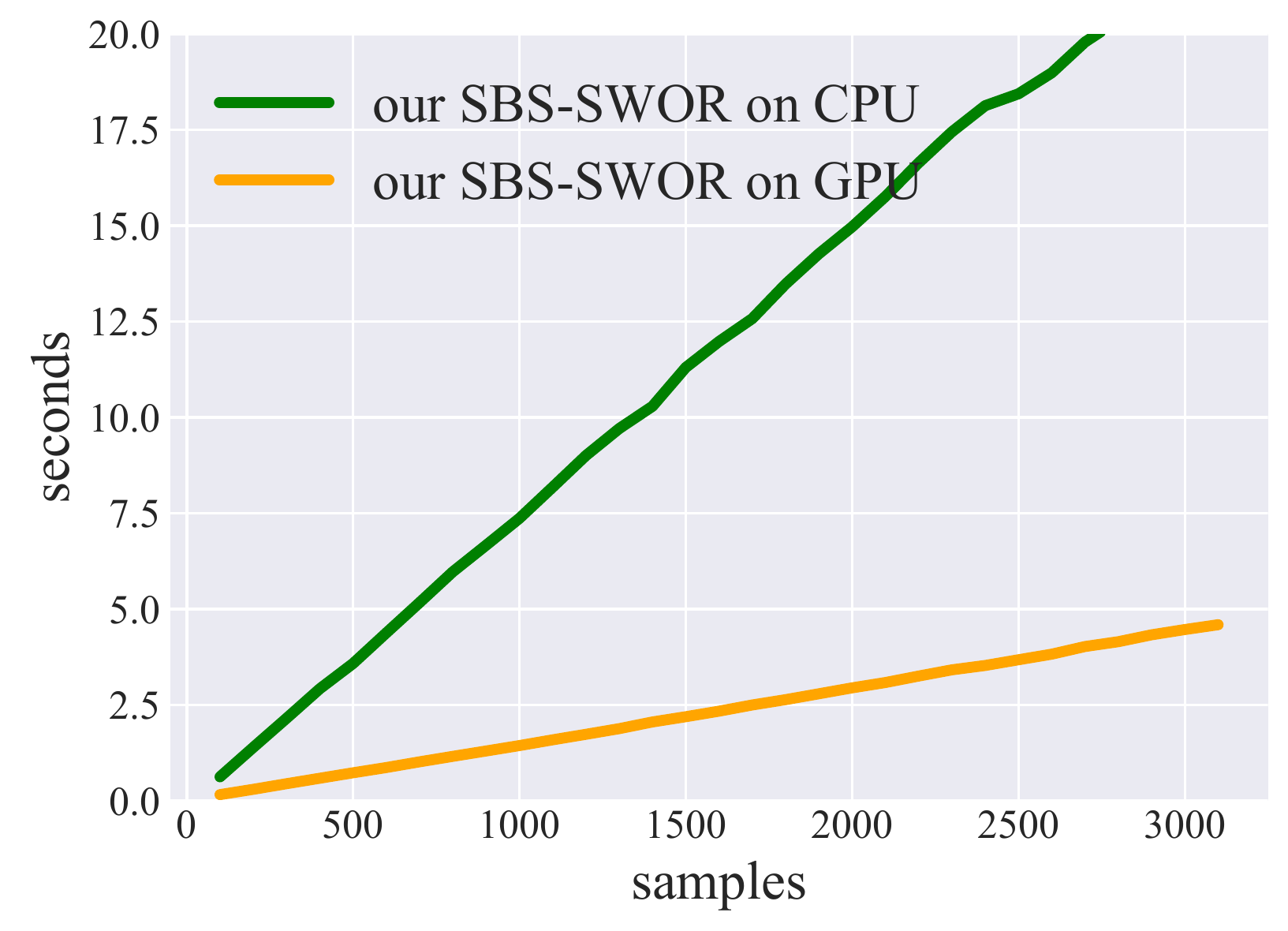}
      \caption{SBS-SWOR in JAX with $100$ words.}
      \label{fig:gpu:speed:simple}
  \end{figure}
}

\newcommand{\trickyGraphFigure}{
    \begin{figure*}[h!]
      \centering
      \subfloat[][Example graph]{
        \scalebox{0.5}{
        \begin{tikzpicture}[->,>=stealth',shorten >=1pt,auto,node distance=2.8cm, semithick]
          \tikzstyle{every state}=[fill=brickred,draw=none,text=white]
    
          \node[state]         (A)                    {$A$};
          \node[state]         (R) [above right of=A] {$R$};
          \node[state]         (B) [below right of=A] {$B$};
          \node[state]         (C) [below right of=R] {$C$};
    
          \path (A) edge [bend left]  node[scale=1.3] {0.5} (C)
                    edge [bend left]  node[scale=1.3] {0.5} (B)
                (R) edge [bend right] node[scale=1.3,above,anchor=south east]{0.5\ \ } (A)
                    edge [bend left]  node[scale=1.3] {0.5} (C)
                (C) edge [bend left]  node[scale=1.3] {0.5} (B)
                (B) edge [bend left]  node[scale=1.3] {0.5} (A);
        \end{tikzpicture}
        }
        \label{fig:single:root:full}
      }\subfloat[][Tree 1]{
        \scalebox{0.5}{
        \begin{tikzpicture}[->,>=stealth',shorten >=1pt,auto,node distance=2.8cm, semithick]
          \tikzstyle{every state}=[fill=brightcerulean,draw=none,text=white]
    
          \node[state]         (A)                    {$A$};
          \node[state]         (R) [above right of=A] {$R$};
          \node[state]         (B) [below right of=A] {$B$};
          \node[state]         (C) [below right of=R] {$C$};
    
          \path (A) edge [bend left]  node[scale=1.3] {0.5} (C)
                (R) edge [bend right] node[scale=1.3,above,anchor=south east]{0.5\ \ } (A)
                (C) edge [bend left]  node[scale=1.3] {0.5} (B);
        \end{tikzpicture}
        }
        \label{fig:single:root:one}
      }\subfloat[][Tree 2]{
        \scalebox{0.5}{
        \begin{tikzpicture}[->,>=stealth',shorten >=1pt,auto,node distance=2.8cm, semithick]
          \tikzstyle{every state}=[fill=brightcerulean,draw=none,text=white]
    
          \node[state]         (A)                    {$A$};
          \node[state]         (R) [above right of=A] {$R$};
          \node[state]         (B) [below right of=A] {$B$};
          \node[state]         (C) [below right of=R] {$C$};
    
          \path (A) edge [bend left]  node[scale=1.3] {0.5} (C)
                    edge [bend left]  node[scale=1.3] {0.5} (B)
                (R) edge [bend right] node[scale=1.3,above,anchor=south east]{0.5\ \ } (A);
        \end{tikzpicture}
        }
        \label{fig:single:root:two}
      }\subfloat[][Tree 3]{
        \scalebox{0.5}{
        \begin{tikzpicture}[->,>=stealth',shorten >=1pt,auto,node distance=2.8cm, semithick]
          \tikzstyle{every state}=[fill=brightcerulean,draw=none,text=white]
    
          \node[state]         (A)                    {$A$};
          \node[state]         (R) [above right of=A] {$R$};
          \node[state]         (B) [below right of=A] {$B$};
          \node[state]         (C) [below right of=R] {$C$};
    
          \path (R) edge [bend left]  node[scale=1.3] {0.5} (C)
                (C) edge [bend left]  node[scale=1.3] {0.5} (B)
                (B) edge [bend left]  node[scale=1.3] {0.5} (A);
        \end{tikzpicture}
        }
        \label{fig:single:root:three}
      }
      \caption{Example graph and its single-root trees where \citeauthor{zmigrod:sampling} algorithm fails to provide unbiased samples.}
      \label{fig:single:root}
    \end{figure*}
}

\newcommand{\algoWilsonRejection}{
    \begin{algorithm}[t]
      \caption{\textproc{WilsonReject}}
      \label{alg:wilson:reject}
      \begin{algorithmic}[1]
        \State \algorithmicdo\ \ $t \gets \Call{Wilson}{G}$
        \State \algorithmicuntil\ \ $t$ has only one ROOT edge
        \State $\Return\ \ t$
      \end{algorithmic}
    \end{algorithm}
}

\newcommand{\algoWilsonMarginal}{
    \begin{algorithm}[t]
      \caption{\textproc{WilsonMarginal}}
      \label{alg:wilson:marginal}
      \begin{algorithmic}[1]
        \State $M \gets \Call{marginals}{G}$                                
        \State $e \gets \Call{sample}{\text{ROOT edges by }M}$              
        \State $G' \gets G \setminus \{\text{all ROOT edges}\} \cup \{e\}$  
        \State $\Return\ \ \Call{Wilson}{G'}$                               
      \end{algorithmic}
    \end{algorithm}
}

\newcommand{\weight}{\operatorname{\phi}}

\newcommand{\algoColbournSimplified}{
    \begin{algorithm}[t]
      \caption{\textproc{Colbourn}'s sampling algorithm.}
      \label{alg:colbourn:simple}
      \begin{algorithmic}[1]
        \State $t \gets \emptyset$ \Comment{Sampled tree edges where\ \ \ \ \ \ \ \ \ }
        \Statex \Comment{$t[i] = j$ stands for edge $j\rightarrow i$}
        \State \label{line:colbourn:marginal} $M \gets \text{marginals}(W) $
        \For{$i \in [1 \dots n]$} \Comment{Loop over words}
          \State \label{line:colbourn:sample} Sample node $v$ with weight $M({v \rightarrow u})$
          \State $t[i] \gets v$
          \State \label{line:colbourn:constrain} $M \gets \text{constrain}(M, v, i)$
        \EndFor
        \State $\Return\ t$
      \end{algorithmic}
    \end{algorithm}
}

\newcommand{\algoWilson}{
    \begin{algorithm}[t]
      \caption{\textproc{Wilson}'s sampling algorithm.}
      \label{alg:wilson}
      \begin{algorithmic}[1]
        \State $t \gets \emptyset$ \Comment{Sampled tree edges where\ \ \ \ \ \ \ \ \ }
        \Statex \Comment{$t[i] = j$ stands for edge $j\rightarrow i$}
        \State \text{visited} $\gets \{\text{ROOT}\}$
        \For{$i \in [1 \dots n]$} \Comment{Loop over words}
          \State $u \gets i$
          \While{$u \notin \text{visited}$} \label{wilson:first:start:ln}
            \State Sample node $v$ with weight $\weight({v \rightarrow u})$
            \State $t[u] \gets v$
            \State $u \gets v$
          \EndWhile \label{wilson:first:end:ln}
          \State $u \gets i$
          \While{$u \notin \text{visited}$} \label{wilson:second:start:ln}
            \State visited.add($u$)
            \State $u \gets t[u]$
          \EndWhile \label{wilson:second:end:ln}
        \EndFor
        \State $\Return\ t$
      \end{algorithmic}
    \end{algorithm}
}

\title{Unbiased and Efficient Sampling of Dependency Trees}
\author{Milo\v{s} Stanojevi\'{c} \\ DeepMind \\ \texttt{stanojevic@deepmind.com}}

\begin{document}

\maketitle

\thispagestyle{plain}

\begin{abstract}

Most computational models of dependency syntax consist of distributions over spanning trees. However, the majority of dependency treebanks require that every valid dependency tree has a single edge coming out of the ROOT node, a constraint that is not part of the definition of spanning trees. For this reason all standard inference algorithms for spanning trees are sub-optimal for inference over dependency trees.

\citet{zmigrod:sampling} proposed algorithms for sampling with and without replacement from the dependency tree distribution that incorporate the single-root constraint. In this paper we show that their fastest algorithm for sampling with replacement, Wilson-RC, is in fact producing biased samples and we provide two alternatives that are unbiased. Additionally, we propose two algorithms (one incremental, one parallel) that reduce the asymptotic runtime of algorithm for sampling $k$ trees without replacement to $\mathcal{O}(kn^3)$. These algorithms are both asymptotically and practically more efficient.

\end{abstract}

\section{Introduction}

Dependency trees are one of the core structures for representing syntactic relations between words \citep{kubler2009dependency,Osborne:book}.
There are multiple ways that these trees can be formally defined which has computational, linguistic and modeling consequences.
Until recently, the most common formalisation of dependency structures used in computational linguistics was a \textit{spanning tree} formalisation. A spanning tree is a directed graph that contains $n+1$ nodes ($n$ words and one artificial ROOT token), $n$ directed edges (each word has one incoming edge except for the ROOT node that has only outgoing edges) and there are no cycles. This simple formalisation allows for the usage of efficient algorithms for finding optimal spanning trees \citep{tarjan77,mcdonald2005mst}.

Unfortunately, spanning trees are too loose of a formalisation to represent the constraints linguists would expect from dependency trees. One of the main linguistic constraints is that there is only one word that represents the main head of the sentence. This is a standard in Universal Dependencies treebanks \citep{ud2-paper} where the constraint is incorporated by stating that every valid dependency tree should have exactly one edge coming out of the artificial ROOT node \citep{udroot}.\footnote{There are few treebanks that make exemption to this rule, e.g. Prague Dependency Treebank \citep{Bejcek2013}.}
In the rest of the article we will use the term \emph{spanning tree} to refer to the unconstrained trees and term \emph{dependency tree} to refer to the trees that have a single-root constraint as in Universal Dependencies.

The distinction between spanning and dependency trees at first may appear unimportant because all dependency trees are a subset of the set of spanning trees and maybe a strong neural model would learn distributions that ignore spanning trees which are not valid dependency trees.
However, \citet{zmigrod:please:mind:the:root} have shown that that is not the case---state-of-the-art trained parsers often predict an invalid dependency tree as the mode and the situation is even worse with k-best trees \citep{zmigrod:k:best}.
This motivates the creation of specialized algorithms that do inference \emph{only} over valid dependency structures.

The most important algorithms needed for inference are: (1) finding the most probable tree, (2) finding marginals over edges, and (3) generating tree samples.
Finding the most probable dependency tree is a solved problem since the algorithm by \citet{stanojevic:cohen} is optimal for that task. Marginals can be computed efficiently using the adaptation of Matrix-Tree Theorem to single-root dependency trees by \citet{koo-etal-2007-structured}. 

In this paper we focus on the last of these problems---generation of samples of dependency trees.
The sampling of dependency trees is often used as a key component of unsupervised grammar induction \citep{marecek-zabokrtsky-2011-gibbs}, semi-supervised training of parsers \citep{corro2018differentiable}, or for enabling the approximate decoding of higher-order models \citep{zhang-etal-2014-greed}.

The only previous work that deals specifically with sampling from distributions of constrained dependency trees is by \citet{zmigrod:sampling}. They adapt algorithms for sampling spanning trees to include the single-root constraint and provide algorithms for sampling with and without replacement. 

Our main contributions are in: (i) showing that the most efficient sampling algorithm by \citeauthor{zmigrod:sampling} is in fact producing biased samples, (ii) proposing two new algorithms that address this issue, and (iii) providing two more algorithms (one incremental and one parallel) that improve the worst-case complexity of sampling without replacement (SWOR). All these algorithms are shown to be both asymptotically and practically efficient.

\section{Distributions over Dependency Trees}

Here we define more formally the distributions over dependency and spanning trees. We will consider only unlabelled trees, but extending this to the labelled case is trivial.

Dependency graph $G$ of a sentence with $n$ words is an unlabelled directed weighted rooted graph that consists of:
\begin{description}[noitemsep,topsep=1pt]
    \item[$n\!+\!1$ nodes] where each word of the sentence is a node, and in addition to those $n$ nodes there is a special ROOT node,
    \item[$n^2$ edges] out of which $n(n-1)$ are directed edges between every pair of word nodes (excluding self-loops) and $n$ edges that go from the special ROOT node to each of the word nodes,
    \item[$n^2$ weights]\!\!, sometimes also called potentials, are non-negative real numbers associated with each of the edges using a function $\weight(\cdot)$.%
    \footnote{The type of models in which a weight can be associated only with a single edge are called \emph{first-order models}. For a comment on higher-order models see the limitations section.
    }
\end{description}

We can represent any graph of this type with an adjacency matrix $W$ of shape $(n\!+\!1, n\!+\!1)$ where all elements are non-negative weights with a constraint that diagonal entries are $0$ (self-loops are not allowed) and first column entries are $0$ (no edges can enter the ROOT node which is at position $0$ by convention).

\textbf{Spanning tree} is any subgraph of $G$ for which it holds that (i) every word node has exactly one incoming edge and (ii) there are no cycles.

\textbf{Dependency tree} is spanning tree that in addition to the above mentioned properties also contains the constraint that there can be only one edge coming out of the special ROOT node.

We will refer to the set of all spanning trees as $T$ and set of all dependency trees as $D$. The following definitions are presented for dependency trees but their analogs for spanning trees are trivial to define by just replacing every occurrence of $D$ with $T$.

\textbf{Weight of a dependency tree} $t$ is a product of the weights of all its edges:
\begin{equation}
    \weight(t) = \prod_{e \in t} \weight(e) \nonumber
\end{equation}

\textbf{Probability of a dependency tree} $t$ is a normalized weight of a tree $t$ with respect to all other dependency trees:
\begin{equation}
    p(t) = \frac{\weight(t)}{\sum_{t' \in D} \weight(t')} = \frac{\weight(t)}{Z_D} \nonumber
\end{equation}

Here $Z_D$ stands for the partition function of dependency trees that must be positive i.e. $Z_D > 0$.

\textbf{Marginal probability} of an edge $e$ in the distribution over dependency trees is a sum of probabilities of all dependency trees $D_e$ that contain edge $e$:
\begin{equation}
    p(e) = \sum_{t \in D_e} p(t) = \frac{\sum_{t \in D_e} \weight(t)}{\sum_{t' \in D} \weight(t')} \nonumber
\end{equation}

Sampler of dependency trees is \textbf{unbiased} \emph{iff} it independently draws random dependency trees with the probability proportional to their weight, i.e. probability of sampling any tree $t$ is $p(t)$.

\section{Previous Work on Sampling Spanning Trees}

Here we present the main two algorithms for sampling from the distributions of unconstrained spanning trees. Both of these algorithms have different advantages. The first one, by \citet{wilson}, is oftentimes faster in practice and we will use it to design a fast sampler of dependency trees in Section~\ref{sec:sampling:with:replacement}. The second algorithm is a type of ancestral sampling by \citet{colbourn} and has a convenient form for extension to sampling without replacement, that will be used in Section~\ref{sec:sampling:without:replacement}.

\subsection{\textproc{Wilson} Sampler}

\citeauthor{wilson}'s algorithm for sampling unconstrained directed spanning trees is in practice the fastest algorithm for that task.
It runs in $\mathcal{O}(h)$ expected computational complexity where $h$ is the \emph{mean hitting time}.
Mean hitting time depends on the weights, but it can often be as small as $\mathcal{O}(n)$ \citep{wilson}.

\algoWilson{}

We use Wilson's algorithm as a black box, so understanding the details of its workings is not important, but we provide it in Algorithm~\ref{alg:wilson} for completeness. Each iteration of the for loop starts from some word (i.e. non-ROOT node) that is not visited and walks a random chain until it reaches some node that was visited in the previous iterations of the for loop. This chain will be attached to the tree that is being sampled. The random walk may contain cycles which are implicitly deleted by overriding parent pointers in lines 7 and 8. The second while loop just marks the nodes in the random walk as visited. In the end of this function all the word nodes will be visited, have one incoming arc, and there will be no cycles.

\subsection{\textproc{Colbourn} Sampler}
\label{sec:colbourn:explanation}

\citet{colbourn} constructed a sampling algorithm that is very different in nature to the \citeauthor{wilson} in that it is an ancestral sampling method that has a fixed runtime complexity that depends only on the size of the graph, but not on the weights. If we have a sentence with words $w_1 w_2 \dots w_n$ we can represent any dependency tree over it by having a sequence of integers $t_1 t_2 \dots t_n$ where the value of $t_i$ defines an edge $w_{t_i} \rightarrow w_i$. Colbourn's algorithm samples each of these integers one by one. After sampling each integer (i.e. edge) we need to constrain the weights of the graph so that the probability of sampling future edges is conditioned on the previously sampled edges. The underspecified version of this algorithm is shown in Algorithm~\ref{alg:colbourn:simple}. A na\"ive implementation of this algorithm could be done in $\mathcal{O}(n^4)$ because we can recompute marginals $n$ times and recomputing marginals each time from scratch takes $\mathcal{O}(n^3)$. This cubic complexity bottleneck comes from the matrix-inverse that is used by Matrix-Tree Theorem to compute marginals \citep{koo-etal-2007-structured}. The main contribution of \citeauthor{colbourn} is a clever way of reducing the complexity of this bottleneck down to $\mathcal{O}(n^2)$ by keeping around the running version of the matrix-inverse and then updating it for each new changed column using Sherman-Morrison formula for rank-one update of matrix-inverse \citep{sherman:morrison}. The technical details of this formula are not important for this presentation, but for completeness can be found in Appendix~\switchCameraReady{sec:appendix:colbourn}{B}. With this trick the total worst-case complexity is $\mathcal{O}(n^3)$. 

\algoColbournSimplified{}

\trickyGraphFigure{}

\citet{zmigrod:sampling} showed that even though Colbourn's algorithm was originally designed for unconstrained spanning trees it can easily be extended to sampling dependency trees by just using the version of Matrix-Tree Theorem by \citet{koo-etal-2007-structured}.
Appendix~\switchCameraReady{sec:appendix:MTT}{A} contains details on MTT.

\section{Extensions of Wilson's Algorithm to Valid Dependency Trees}
\label{sec:sampling:with:replacement}

In this section we will describe novel contributions with regards to sampling dependency trees with replacement. In \S\ref{sec:zmigrod:biased} we show that the previous fastest sampling algorithm is in fact biased. In \S\ref{sec:wilson:marginal} and \S\ref{sec:wilson:reject} we present two new unbiased algorithms.

\subsection{\textproc{WilsonRC} Extension by \citeauthor{zmigrod:sampling} is Biased}
\label{sec:zmigrod:biased}

\citet{zmigrod:sampling} propose a very simple extension of the Wilson's algorithm to the single-root dependency trees that works in following steps:
\begin{enumerate}[itemsep=0pt,topsep=5pt]
    \item sample an edge $e$ from the set of all edges that are coming out of ROOT by using their original weights $\phi(e)$,
    \item construct graph $G'$ by deleting all edges coming out of ROOT
    and set the target of the edge $e$ to be the new root,
    \item apply original Wilson's algorithm to graph $G'$.
\end{enumerate}

This extension is very simple, but unfortunately, as we show here, it is biased.%
\footnote{Since the publication of our paper \citet{zmigrod:sampling} have modified their paper to acknowledge that \textproc{WilsonRC} is biased. \citet{zmigrod:thesis} PhD thesis is an excellent work containing many important algorithms for non-projective dependency distributions. It also includes re-implementation and additional benchmarks of our unbiased algorithms.}
We demonstrate that that \citet{zmigrod:sampling} algorithm is biased with a simple example graph in Figure~\ref{fig:single:root:full}.
For a formal proof see Appendix~\ref{sec:appendix:wilsonrc:biased:proof}.

This graph has only three constrained dependency trees that are shown in Figures~\ref{fig:single:root:one}-\ref{fig:single:root:three}. They all have same weights so any unbiased sampling algorithm should return them with the same probability. However, the step 1 of \citeauthor{zmigrod:sampling} will half of the time sample ROOT edge $R\rightarrow A$  for the first two trees and half of the time the ROOT edge $R\rightarrow C$ for the last tree and therefore it is biased to sample the last tree over the first two.
In the following sections we propose unbiased alternatives.

\subsection{Improvement 1: WilsonMarginal}
\label{sec:wilson:marginal}

Intuitively, by looking at the example in Figure~\ref{fig:single:root} the first step that samples the single root edge should reflect our intuition that edge $R\rightarrow A$ should be sampled with two times higher probability than edge $R\rightarrow C$ because it has two times more of the (weighted) trees. This can be accomplished by sampling from the \emph{marginal} of each root edge. Marginal for the edges $R\rightarrow A$ and $R\rightarrow C$ in this graph will be $\frac{2}{3}$ and $\frac{1}{3}$ respectively.

\algoWilsonMarginal{}

\textproc{WilsonMarginal} (shown in Algorithm~\ref{alg:wilson:marginal}) does exactly that. Step 1 is computing marginals which can be done using \citet{koo-etal-2007-structured} version of the Matrix-Tree Theorem in $\mathcal{O}(n^3)$. Steps 2 and 3 run in $\mathcal{O}(n)$ since that is the maximal number of edges coming out of ROOT. Step 4 is the standard Wilson's algorithm that runs in $\mathcal{O}(h)$. Therefore getting a single sample with this algorithm can be done in $\mathcal{O}(n^3+h)$. If we want to draw multiple samples from the same graph we can reuse the computation of the marginals. Drawing $k$ samples from the same graph can be done in $\mathcal{O}(n^3+kh)$.

\begin{theorem}
\textproc{WilsonMarginal} is an unbiased sampling algorithm i.e. it samples each dependency tree $t \in D$ with probability:
\begin{equation}
  p(t) = \frac{\weight(t)}{\sum_{t' \in D} \weight(t')} \nonumber
\end{equation}
\end{theorem}

\noindent{}\textit{Proof.}
The root edge of the sampled tree $t$ will be sampled with probability
$\frac{\sum_{t' \in D_e} \weight(t')}{\sum_{t' \in D} \weight(t')}$
where $D_e$ is the subset of all dependency trees that contain edge $e$.
The rest of the tree will be sampled from graph $G'$ with probability
$\frac{\weight(t)}{\sum_{t' \in D_e} \weight(t')}$.
The probability of sampling tree $t$ using \textproc{WilsonMarginal} is 
$\frac{\sum_{t' \in D_e} \weight(t')}{\sum_{t' \in D} \weight(t')} \frac{\weight(t)}{\sum_{t' \in D_e} \weight(t')} = \frac{\weight(t)}{\sum_{t' \in D} \weight(t')}= p(t)$
\hfill{}$\blacksquare{}$

\subsection{Improvement 2: WilsonReject}
\label{sec:wilson:reject}

The complexity of \textproc{WilsonMarginal} is worse than the original unconstrained Wilson's algorithm because of the cubic term.
This term comes from the need to invert the Laplacian matrix during the computation of the marginals. This operation is optimized in many software packages \citep{fast:matrix:inverse} and in practice it does not take a long time to compute.%
 Still, it would be good to avoid computing matrix-inverse unless necessary because matrix-inverse can be numerical unstable if the matrix is ill-conditioned \citep{matrix:inverse:instability,matrix:inverse:instability:book}.
Here we present an alternative based on rejection sampling that on average works well for most graphs that are of interest.

\newcommand{\M}{c}

Rejection sampling is used in cases when it is difficult to sample from the distribution of interest $p(t)$ but it is easy to sample from some related proposal distribution $q(t)$.
The proposal distribution $q(t)$ needs to satisfy $\M{} q(t) \geq \tilde{p}(t)$ for some constant $\M{}>0$ where $\tilde{p}(t)$ is the unnormalized target distribution. In other words, $\M{} q(t)$ needs to form an upper envelope of $\tilde{p}(t)$. After sample is retrieved from $q(t)$ it is accepted as a sample of $p(t)$ with probability $\frac{\tilde{p}(t)}{\M{} q(t)}$. See \citet[\S23.3]{Murphy:2012} for a good introduction to rejection sampling.

Barring the envelope condition, we have a complete freedom to choose the proposal distribution $q(\cdot)$ and constant $\M{}$. Any choice that satisfies the envelope condition gives an unbiased sampler.

Here the target distribution from which we want to sample is the distribution over dependency trees $p(t)$. A convenient distribution to choose as a proposal $q(t)$ is the distribution over spanning trees in the same weighted graph. All valid dependency trees $t$ will be present in the support of both $p(\cdot)$ and $q(\cdot)$ and their \emph{unnormalized} weight will be the same since they came from the same weighted graph. If a tree is a spanning tree $t'$ that has more than one root edge then it will be in the support of $q(\cdot)$ but not of $p(\cdot)$ so their unnormalized score for target distribution is $\tilde{p}(t')=0$.  More formally:
\begin{align}
    \tilde{p}(t) =
    \begin{cases}
        \tilde{q}(t), & \text{if\ \ $t \in D$}\\
        0,& \text{otherwise}
    \end{cases} \label{eq:unnormalized:probs}
\end{align}

In this setup the unnormalized $\tilde{q}(t)$ forms an envelope over unnormalized $\tilde{p}(t)$ because $\forall t. \tilde{q}(t)\geq \tilde{p}(t)$. Still, this is not sufficient to apply rejection sampling because we need an envelope with the \emph{normalized} distribution $q(t)$. To get that we multiply the normalized proposal with constant $\M{}=Z_T$ which is the partition function over the set of unconstrained spanning trees.

\begin{theorem}
\textproc{WilsonReject} is an unbiased sampling procedure for dependency tree distributions.
\end{theorem}

\noindent{}\textit{Proof.}
Given that we already have an unbiased way for generating proposal samples from $q(\cdot)$, i.e. the original Wilson algorithm that generates spanning trees, we only need to prove that the chosen proposal $q(\cdot)$ and constant $\M{}$ satisfy the envelope condition for rejection sampling $\M{} q(t) \geq \tilde{p}(t)$:
    $$\M{} q(t) = Z_T \frac{\tilde{q}(t)}{Z_T} = \tilde{q}(t) \geq \tilde{p}(t)$$
The last step follows from Equation~\ref{eq:unnormalized:probs}.
\hfill{}$\blacksquare{}$

When the rejection sampling algorithm gets a sample from $q(\cdot)$ it accepts it with probability:
\begin{align}
  \frac{\tilde{p}(t)}{\M{} q(t)} = \frac{\tilde{p}(t)}{Z_T \frac{\tilde{q}(t)}{Z_T}} = 
    \begin{cases}
        1,& \text{if\ \ $t \in D$}\\
        0,& \text{otherwise}
    \end{cases} \nonumber
\end{align}

This gives us a very simple implementation of the \textproc{WilsonReject} sampling algorithm presented in Algorithm~\ref{alg:wilson:reject}.
This algorithm samples spanning trees with the original Wilson's algorithm until it gets a valid dependency tree.

\algoWilsonRejection{}

\subsubsection{Expected number of runs for WilsonReject}
\label{sec:expected:runs:wilson:reject}

The efficiency of \textsc{WilsonReject} depends on how many samples we need to draw from $q(t)$ until we get a valid dependency tree.
We can model the process of getting a successful sample as a geometric distribution in which the probability of success is $\sum_{t \in D} q(t)$. The expected number of draws before success in a geometric distribution is $\frac{1}{\sum_{t \in D} q(t)} = \frac{Z_T}{\sum_{t \in D} \tilde{q}(t)} = \frac{Z_T}{Z_D}$, i.e. the expected number of attempts until success is the ratio of partitions for spanning and dependency trees.

To determine this number we need to know something about actual distribution over trees. In dependency parsing we usually care about two scenarios. The first one is the random weights setting that usually happens in the beginning of training of dependency parsers. The second one is the setting of already trained parsers. We answer the first one formally and second one empirically.

\paragraph{Random weights setting} Here edge weights are sampled independently from the same distribution with mean $\mu$. We are interested in finding expected number of runs under the distribution of graph weights $\expectation{\frac{Z_T}{Z_D}}$. Let $w_{avg}^T$ and $w_{avg}^D$ be the average weights of spanning and dependency trees in a given weighted graph. We can write the following:

\begin{equation}
    \expectation{\frac{Z_T}{Z_D}} = \expectation{\frac{|T|w_{avg}^T}{|D|w_{avg}^D}}  = \frac{|T|}{|D|}\expectation{\frac{w_{avg}^T}{w_{avg}^D}} \label{eq:expected:runs:one}
\end{equation}

The number of spanning trees $|T|$ is given by \citeauthor{cayley1889theorem}'s formula:%
\footnote{This is Cayley’s formula with an offset of $1$ to account for the artificial ROOT node.
Cayley’s formula was originally defined for undirected spanning trees, but it applies equally to \emph{rooted} directed spanning trees.}
\begin{align}
    |T| = C_n = (n+1)^{n-1} \label{eq:count:spanning:trees}
\end{align}

The number of dependency trees $|D|$ is $nC_{n-1}$ because we can pick any of the $n$ words to be the root of the smaller sub-graph of $n-1$ words.
\begin{align}
    |D| = nC_{n-1} = n^{n-1} \label{eq:count:dependency:trees}
\end{align}

Equations \ref{eq:expected:runs:one}, \ref{eq:count:spanning:trees} and \ref{eq:count:dependency:trees} together give us an upper bound on the expected number of tries before \textproc{WilsonReject} finds a successful sample:

\begin{equation}
    \expectation{\frac{Z_T}{Z_D}} = \frac{(n+1)^{n-1}}{n^{n-1}}\expectation{\frac{w_{avg}^T}{w_{avg}^D}} 
    < e \expectation{\frac{w_{avg}^T}{w_{avg}^D}} \nonumber
\end{equation}

The expectation of averages is difficult to solve analytically, as usual with ratio distributions, because the weights of spanning and dependency trees in a given weighted graph are not independent. If we assume that average weights of spanning and dependency trees in a given weighted graph are approximately the same, which is a reasonable assumption since their expected weights across all possible weightings are the same, the solution to inequality above will be $e$.

To verify how reasonable is this assumption we have ran a large number of simulations on different graph sizes and weight distributions over edges. We found that $w_{avg}^D \approx w_{avg}^T$ to be very accurate in all settings we have tested.
Appendix~\switchCameraReady{sec:appendix:assumption}{D} contains the results of the simulations.

All this taken together means that on average \textproc{WilsonReject} will need to sample approximately less than three times before it finds a successful sample in a randomly weighted graph.

\paragraph{Trained weights setting}
In this setting we would expect that the model that was trained only on single-root trees to automatically put most of the probability mass on single-root trees, i.e. $Z_D \lessapprox Z_T$. To test this we used the model of Stanza parser \citep{stanza} trained for English. We take English as the most extreme example since it has more training data than other languages and \citet{zmigrod:please:mind:the:root} show that that makes the model more likely to put the mode of the distribution on a valid dependency tree. We would expect that models for languages with less training data would be somewhere between the random weights setting and English trained weights setting. We applied the parser to the English portion of the News Commentary v16 corpus
and computed $Z_D$ and $Z_T$ exactly using the algorithm of \citet{koo-etal-2007-structured}. On almost all sentences the Stanza model puts more than $95\%$ of probability mass on the single-root trees. This means that the average number of tries needed by \textproc{WilsonReject} to get a successful sample on a trained graph is $\approx 1.06$.

The average behavior of \textproc{WilsonReject} on both random weights and trained weights setting is as good as the unconstrained original Wilson's sampler except for a small multiplicative constant.

\section{Sampling Without Replacement (SWOR)}
\label{sec:sampling:without:replacement}

Sampling without replacement (SWOR) is a sampling procedure where after some instance is sampled it cannot be sampled again. That is useful in cases of low-entropy distributions where standard sampling is likely to provide many repetitions of the same set of samples and be inefficient in estimating the target value. Samples are not independently drawn and therefore require specialized estimators \citep{vieira2017estimating,Kool2020Estimating}.

\citet{zmigrod:sampling} propose a SWOR algorithm that is based on modification of Colbourn's ancestral sampling algorithm.
They follow the general pattern for constructing SWOR algorithms by \citet{shi2020incremental} where after one tree is sampled, its probability mass is removed from the distribution so it cannot be sampled again. \citeauthor{zmigrod:sampling} maintain an unstructured list of sampled trees that is always queried whenever a new sample is drawn. This makes the algorithm have quadratic complexity in terms of the number of drawn samples. Concretely, for a sentence of length $n$ the SWOR algorithm of \citeauthor{zmigrod:sampling} draws $k$ samples in $\mathcal{O}(kn^3+k^2n)$. This is inefficient if we want to draw a large number of samples.

We provide two solutions that reduce complexity to $\mathcal{O}(kn^3)$ which is linearly dependent on the number of samples $k$. The main idea for both algorithms is to represent distributions over dependency trees as sequential auto-regressive models. After that is done we can apply any algorithm for sampling without replacement from sequential models.

\subsection{Dependency Tree Distributions in Sequential Auto-Regressive Form}
\label{sec:tree:dist:in:autoregressive}

Colbourn's algorithm from Section~\ref{sec:colbourn:explanation} can be interpreted as an auto-regressive form of dependency tree distributions.
Here we restate presentation from Algorithm~\ref{alg:colbourn:simple} as a state machine in which:
\begin{description}[noitemsep,topsep=1pt]
    \item[state] is a sequence of edges sampled until this point, together with additional information needed for efficient computation of marginals such as an inverse of the graph's Laplacian,
    \item[transition] from any state to another state adds one more edge that enters the upcoming word, i.e. the upcoming word's head is determined.
\end{description}

The initial state contains an empty set of edges and has transitions that generate the incoming edge for the first word in the sentence. The probabilities of these initial transitions are computed using a method of computing marginals with Matrix-Tree Theorem in $\mathcal{O}(n^3)$ \citep{koo-etal-2007-structured} as in line~\ref{line:colbourn:marginal} of Algorithm~\ref{alg:colbourn:simple}. This computation is done only once because it is needed only for the initial state. When the state machine transitions from one state to another it selects one of the dependency arcs and then constrains the graph (and its marginals) so that any upcoming selection of incoming edges for future words will have to condition on the newly selected edge. This can be done efficiently using Sherman-Morrison formula in $\mathcal{O}(n^2)$ \citep{sherman:morrison}. The detailed formulation of the state machine, pseudo-code and usage of Sherman-Morrison formula is presented in Appendix~\switchCameraReady{sec:appendix:colbourn}{B}.
This state-machine needs to take $n$ transitions to go from initial state to the final state where each transition costs $\mathcal{O}(n^2)$. In total, unfolding one complete transition sequence takes $\mathcal{O}(n^3)$ which is expected since this is just a reformulation of Colbourn's algorithm.

Now that we have Colbourn's algorithm as an auto-regressive model we can use any SWOR algorithm for sequential models. Two popular options are Trie algorithm by \citet{shi2020incremental} and Stochastic Beam Search (SBS) by \citet{stochastic:beams}.

\citeauthor{shi2020incremental}'s Trie algorithm for sampling sequences uses an efficient trie data-structure to remove previously sampled sequences from the support of the probability distribution so that the sequence that will be sampled next will not be from the set of already sampled sequences.

\citeauthor{stochastic:beams}'s Stochastic Beam Search takes a different approach that is based on the generalization of Gumbel trick. \citet{gumbel1954statistical} has shown that sampling from a categorical distribution can be reformulated as finding an \textit{argmax} from logits with special noise added to them. \citet{vieira2014gumbel} has shown that if \textit{argmax} is replaced by \textit{top-k} what we get is $k$ samples without replacement for categorical distribution. \citet{stochastic:beams,kool2020ancestral} generalized this to structured sequential distributions by using beam search and carefully controlling the injected noise.

The worst-case complexity of getting $k$ SWOR samples from each of these algorithms is $\mathcal{O}(knt)$ where $t$ is the complexity of taking a single transition. With our representation of Colbourn's algorithm each transition takes $\mathcal{O}(n^2)$ which means the total worst-case complexity is $\mathcal{O}(kn^3)$.

We refer the reader to the papers of \citet{shi2020incremental} and \citet{stochastic:beams} for the details of Trie SWOR sampling and SBS since the details are not of particular importance for our approach, in principle any ancestral SWOR could be applied in our case, and we cannot do justice to those works by presenting them briefly. We focus instead on how our approach differs from previously proposed one by \citet{zmigrod:sampling}. Given that both our Trie-SWOR and the SWOR of \citeauthor{zmigrod:sampling} are based on extending \citet{shi2020incremental} with Colbourn's algorithm it may seem strange that they get a worse worst-case complexity of $\mathcal{O}(kn^3+k^2n)$. That is because \citeauthor{zmigrod:sampling} maintain previous samples in an unstructured list of size $\mathcal{O}(k)$ so removing the probability of already sampled sequences requires an additional pass over this list for each new sample. We instead follow \citet{shi2020incremental} in using a trie data structure and do not need to make any additional passes.

Having both Trie and SBS SWOR options is not redundant as each of these approaches have pros and cons. The advantage of Trie algorithm is that it draws samples incrementally one by one so we can dynamically decide when we have a sufficient number of samples. The advantage of Stochastic Beam Search is that the samples are drawn in parallel which is more efficient on hardware like GPUs.

\section{Experiments}

\figureNormalSpeed{}

We test the empirical performance of the algorithms presented here. We implemented all of them in Python and NumPy in order to comparable to previous work on \citet{zmigrod:sampling} for SWOR sampling. We do not compare performance against their adaptation of Wilson's algorithm for single-root trees since that adaptation is not correct, as we have shown in Section~\ref{sec:zmigrod:biased}.

\paragraph{Sampling with Replacement} experiments were conducted in two settings: random weights and trained weights.
For random weights we used graphs with weights sampled from the uniform distribution.
To get the trained weights we applied Stanza's English model \citep{stanza} on English sentences of the News Commentary corpus.

For each graph we sample $100$ trees with replacement. We compared \textproc{Colbourn} with single-root constraint from \citet{zmigrod:sampling} against our \textproc{WilsonMarginal} and \textproc{WilsonReject} from Section~\ref{sec:sampling:with:replacement}.
The results are shown in Figure~\ref{fig:normal:speed}.

Both of our algorithms significantly improve upon the version of Colbourn's algorithm presented in \citet{zmigrod:sampling}.
Among the two of our algorithms, \textproc{WilsonMarginal} performs better in random weights setting, while in trained weights setting their performance is almost identical. This is expected result from the analysis in Section~\ref{sec:expected:runs:wilson:reject}.

In trained weights there are almost no rejected samples so the difference between \textproc{WilsonReject} and \textproc{WilsonMarginal} comes down to \textproc{WilsonMarginal} having matrix-inversion operation which does not seem to cause any slowdown.

\figureSworSpeed{}

\paragraph{Sampling without Replacement} tests were conducted in a similar way except that we vary the number of SWOR samples instead of the sentence length. We use sentence length of $14$ words because that is the longest sequence that we could run \citeauthor{zmigrod:sampling} algorithm without encountering issues with numerical instability. In Appendix~\switchCameraReady{sec:appendix:swor:results}{C} we show results with running our algorithm on the longer sentences. We test only on random weights since the runtime of these algorithms does not depend on the weights value. As visible from Figure~\ref{fig:swor:speed} both of our algorithms outperform the algorithm of \citet{zmigrod:sampling}. Of our two algorithms, the Trie algorithm is slightly slower due to the constant factors needed to maintain the trie data structure.

We also reimplemented a version of SBS-SWOR in JAX so that we can see the benefits of having a parallel algorithm executed with parallel hardware and got up to $\times 5$ speedup on GPU over CPU with sentences of length $100$ (Figure~\ref{fig:gpu:speed:simple}).
See Appendix~\switchCameraReady{sec:appendix:swor:results}{C} for tests on other lengths.
The main takeaway is that SBS-SWOR allows for a vectorized implementation that can exploit modern hardware accelerators, unlike other SWOR algorithms.

\figureGPUSpeedSimple{}

\section{Conclusion}

We have presented multiple contributions to the techniques for sampling dependency trees that (1) correct errors in the previously published algorithms, (2) propose new algorithms for sampling with replacement, (3) propose new algorithms for sampling without replacement and (4) show that these algorithms have both asymptotically and practically good performance. We hope the proposed algorithms will lead to advancements in all problems where sampling is crucial such as unsupervised grammar induction \citep{corro2018differentiable}, syntax marginalization in syntactic language models \citep{transformer:grammars} and approximate decoding of higher-order models \citep{zhang-etal-2014-greed}.

\section*{Limitations}
\label{sec:limitations}

A limitation of our approach is that we have a multitude of different algorithms with their pros and cons instead of one unified algorithm that would have the best properties of all of them.
For instance, Wilson-based algorithms are fast in practice, but they provide only expected runtime and no worst-case complexity. Colbourn-based algorithms have a predictable runtime, but are not as fast in practice.
Ideally we would have an algorithm that has the practical speed of \textsc{WilsonMarginal}, numerical stability of \textsc{WilsonReject}, predictable runtime and flexibility of \textsc{Colbourn}.

One could try to make Colbourn algorithm closer to Wilson's in performance by having a better implementation -- in their paper \citet{colbourn} show how sampling can be reduced to matrix multiplication and therefore use sub-cubic algorithms for implementation such as the ones by \citet{strassen-gaussian-1969}, \citet{coppersmith:winograd:matrix:multiplication} or \citet{AlphaTensor2022}. However, in most cases these sub-cubic algorithms would not reach the speed of Wilson-based algorithms. It would appear that either we have to choose random walk paradigm of Wilson's algorithm or matrix multiplication approach of Colbourn. There are some recent works that show that it may be possible to have a separate paradigm that can with high probability have runtime asymptotically faster than matrix multiplication \citep{durfee2017sampling}.

Another limitation, in comparison to projective dependency parsing and CFG parsing, is that we have a completely separate algorithms for sampling (presented in this paper) and finding an \emph{argmax} (presented by \citet{stanojevic:cohen}) of a non-projective dependency tree distribution. In projective and CFG parsers it is possible to express both algorithms as a single inside algorithm that just uses different semi-rings for \emph{sampling} and \emph{argmax} \citep{goodman-1999-semiring,wilker:grasp,torch-struct}. It is possible to simulate this to some extent by injecting some noise to weights and then finding \emph{argmax} as done by \citet{corro2018differentiable}, but that provides biased samples.

In this work we also do not deal directly with sampling from higher-order models \citep{higher:mcdonald-pereira-2006-online,higher:koo-collins-2010-efficient}. However, the algorithms presented here can be useful for approximate decoding of higher order models. In general, finding the best tree from a higher-order model is NP-complete \citep{mcdonald-satta-2007-complexity}, but as \citet{zhang-etal-2014-greed} show it is possible to decode approximately by sampling from the first-order model and improving the sample using the higher-order model. \citeauthor{zhang-etal-2014-greed} have used the original Wilson's algorithm for sampling spanning trees from uniform distribution which means that on average $\frac{2}{e}\approx 73\%$ of the sampled trees would be invalid dependency trees (see Section~\ref{sec:expected:runs:wilson:reject}). It would be easy to apply any of our sampling algorithms in their setting and get valid dependency trees as a better starting point for decoding a higher-order model.

\section*{Acknowledgements}
I am grateful to Shay Cohen, Clara Meister, G\'{a}bor Melis, Laura Rimell and Chris Dyer for the comments on the earlier versions of this work.

\bibliography{BIB}
\bibliographystyle{acl_natbib}

\clearpage

\appendix

\section{Matrix-Tree Theorem}
\label{sec:appendix:MTT}

Matrix-Tree Theorem (MTT) is an important theorem from graph theory that establishes the connection between spanning trees of a graph and its Laplacian matrix. It allows for an efficient computations of the number of spanning trees present in a graph. It was originally devised for counting undirected spanning trees, but it was later extended to weighted directed spanning trees by \citet{Tutte84}. In $2007$ three different papers simultaneously presented application of MTT to dependency parsing \citep{mcdonald-satta-2007-complexity,smith-smith-2007-probabilistic,koo-etal-2007-structured}. Out of these \citet{koo-etal-2007-structured} is of particular interest here because it extends MTT to have a single-root constraint that all valid dependency trees should have.

As usual, we can represent any weighted directed graph using a weighted adjacency matrix $W$ of shape $(n+1, n+1)$ where $W[i, j]$ defines the weight of an edge $i \rightarrow j$. All weights must be non-negative real numbers. Node $0$ is reserved for the artificial ROOT token while all other nodes are regular words that appear in the sentence. Self-loops are not allowed so all diagonal entries have weight $0$. There cannot be any edges entering ROOT so all entries in the $0$th column are also $0$.

\emph{Degree} matrix $D$ is a diagonal matrix that on the $i$th element of the diagonal contains the weighted in-degree of the $i$th node. The in-degree is simply the sum of weights of all the edges that enter $i$th node.

\emph{Laplacian} matrix $L$ is defined by equation:
\begin{align}
    L = D-W \nonumber
\end{align}

Matrix-Tree Theorem in the form presented by \citet{Tutte84} states that any co-factor of a graph's Laplacian matrix will be the partition function over the distribution of all spanning trees in the graph.
For instance, computing the partition function can be done in the following way:

\begin{align}
  Z = |L[1\!:, 1\!:]| \nonumber
\end{align}

Here we have used Python notation for slicing for convenience. Determinant of this sub-matrix of a Laplacian gives us a partition function. Sometimes in the literature this sub-matrix of a Laplacian is referred to as the Laplacian. The only non-trivial computation here is computation of a determinant that can be done as fast as matrix-multiplication but in most practical cases is implemented in $\mathcal{O}(n^3)$. For convenience let us name the function that computes this sub-matrix $\hat{L}$ i.e.:
\begin{align}
    \hat{L}(W) = (\operatorname{degree}(W)-W)[1\!:, 1\!:]
\end{align}

\citet{koo-etal-2007-structured} make an adjustment to this sub-matrix in order to incorporate the single-root constraint. To do that it is necessary to separately treat the edges that come out of the ROOT node. These edges' weights are located in $W[0, 1\!:]$. The modification by \citeauthor{koo-etal-2007-structured} is (1) to not use ROOT edges in the construction of the degree matrix, and (2) to replace first row of the Laplacian sub-matrix with those edges. Algorithm~\ref{alg:L:hat:appendix} shows a JAX NumPy implementation of the modified $\hat{L}$.

\begin{algorithm}
  \caption{Computation of Laplacian sub-matrix $\hat{L}(W)$ for single-rooted dependency trees.}
  \label{alg:L:hat:appendix}
\begin{lstlisting}[language=Python]
  def L_hat(W):
    Wp = W[1:, 1:]
    I = jnp.eye(Wp.shape[0])
    D = jnp.sum(Wp, axis=0) * I
    L = D - Wp
    return L.at[0].set(Wp[0])
\end{lstlisting}
\end{algorithm}

Note that each line takes $\mathcal{O}(n^2)$ to compute.

Another useful statistic is computation of marginals that can be done by extending MTT further. For finding marginals it often comes handy to use the identity which states that the gradient of a log-partition is equivalent to marginals \citep{wainwright:jordan:2008,eisner-2016-inside}:
\begin{align}
    M(W) = \frac{\partial \log |\hat{L}(W)|}{\partial W} \nonumber
\end{align}

The simplest way to implement this is by using automatic differentiation tools \citep{zmigrod:expectation}, but it can also be done by an explicit computation of a derivative of a determinant which involves inverting Laplacian sub-matrix as presented by \citet{koo-etal-2007-structured}. For the motivation of that derivation see the original paper. In Algorithm~\ref{alg:marginal:given:B:appendix} we present only the few lines of JAX NumPy code that are needed for implementing this computation. The code could be written in a single function but it will become clear in Appendix~\ref{sec:appendix:colbourn} why it is beneficial to have it as two separate functions.

\begin{algorithm}
  \caption{Computation of marginals for single-rooted dependency trees.}
  \label{alg:marginal:given:B:appendix}
\begin{lstlisting}[language=Python]
def marginals(W):
  L = L_hat(W)
  B = jnp.linalg.inv(L).T
  return _for_a_given_B(W, B)
  
def _for_a_given_B(W, B):
  M = jnp.zeros(W.shape)
  X = jnp.diag(B)\
         .at[0].set(0)\
         .reshape(1, -1)
  Y = B.at[0].set(0)
  M = M.at[1:, 1:].set(
          W[1:, 1:] * (X - Y)
      ).at[0 , 1:].set(
          W[0, 1:] * jnp.diag(B)
      )
  return M
\end{lstlisting}
\end{algorithm}

Each line takes $\mathcal{O}(n^2)$ to execute except for the line that computes matrix inversion which takes $\mathcal{O}(n^3)$.

\section{Colbourn Algorithm}
\label{sec:appendix:colbourn}

This section continues on the discussion from Section~\switchCameraReady{sec:colbourn:explanation}{3.2} and Appendix~\ref{sec:appendix:MTT}.
We are repeating the general specification of the Colbourn algorithm in  Algorithm~\ref{alg:colbourn:simple:appendix}.
The Colbourn algorithm samples the edges of a dependency/spanning tree one by one ordered by their end points. So if we have have $n$ words it will first find an edge entering word $w_1$, then edge entering word $w_2$ etc. In the initial state of the algorithm, line~\ref{line:colbourn:marginal:appendix}, the algorithm computes the total probability that is concentrated for each edge that could be chosen for the first word. This is done by simply computing the marginals for the whole graph as in Algorithm~\ref{alg:marginal:given:B:appendix}. Given the marginals it is easy to sample the first edge in line~\ref{line:colbourn:sample:appendix}. After that edge is sampled and stored we need to adjust the marginals so that they condition on the selected edge. A na\"ive way of doing that would be to edit the weight matrix $W$ so that all other edges entering the word have weight $0$ and then rerun Algorithm~\ref{alg:marginal:given:B:appendix} to find the new marginals. If we repeat this process $n$ times we will select one entering edge for each word and form a full dependency tree.

\begin{algorithm}
  \caption{\textproc{Colbourn}'s sampling algorithm.}
  \label{alg:colbourn:simple:appendix}
  \begin{algorithmic}[1]
    \State $t \gets \emptyset$ \Comment{Sampled tree edges where\ \ \ \ \ \ \ \ \ }
    \Statex \Comment{$t[i] = j$ stands for edge $j\rightarrow i$}
    \State \label{line:colbourn:marginal:appendix} $M \gets \text{marginals}(W) $
    \For{$i \in [1 \dots n]$} \Comment{Loop over words}
      \State \label{line:colbourn:sample:appendix} Sample node $v$ with weight $M({v \rightarrow u})$
      \State $t[i] \gets v$
      \State \label{line:colbourn:constrain:appendix} $M \gets \text{constrain}(M, v, i)$
    \EndFor
    \State $\Return\ t$
  \end{algorithmic}
\end{algorithm}

A problem with the na\"ive implementation is that its complexity would be $\mathcal{O}(n^4)$. The main bottleneck is matrix inversion that needs to be done in the computation of marginals after each new edge is added. \citeauthor{colbourn} noticed that it is possible reuse the previously computed matrix-inversion of matrix $W$ to compute the matrix-inversion of the new matrix $W'$ which differs from $W$ by only one column. The method for doing that is Sherman-Morrison formula \citep{sherman:morrison} that makes the computation of marginals alone $\mathcal{O}(n^2)$ and the complexity of the whole Colbourn algorithm down to $\mathcal{O}(n^3)$. Here we present the version that updates $B$ which is not only matrix-inversion but also a transposed matrix. In the $i$th iteration of the loop the algorithm will sample edge $j\rightarrow i$ as the entering edge for the word $w_i$. We can update the matrices with the following equations:
\begin{align}
    W'&[a, b] =
              \begin{cases}
                  0      ,& \text{if } a \neq j \land b = i\\
                  W[a,b] ,& \text{otherwise}
              \end{cases}  \label{eq:sherman:morrison:one:hot}\\
    u & = \hat{L}(W')[:, i] - \hat{L}(W)[:, i]  \label{eq:sherman:morrison:u}\\
    B' & = \frac{B[:, j] (u^T B)}{1+u^T B[:, j]} \label{eq:sherman:morrison:update:B}
\end{align}

Eq~\ref{eq:sherman:morrison:one:hot} will compute the new weight matrix $W'$ by changing only one column of matrix $W$. Eq~\ref{eq:sherman:morrison:u} computes the difference of that column between the Laplacians of old and new weight matrix. Eq~\ref{eq:sherman:morrison:update:B} applies Sherman-Morrison formula to find the updated $B$ matrix.

The main idea of our SWOR sampling algorithms is to use Colbourn's sampling algorithm as a basis for treating distributions over dependency trees as an sequential auto-regressive model and then apply existing sampling algorithms for sequential models to dependency trees. This reformulation is presented in Algorithm~\ref{alg:state:transitions}. This is the same Colbourn algorithm but expressed as a state machine. The state contains the current position, current weight matrix, current Laplacian sub-matrix, and its inverse-transpose. Computation of the initial state takes $\mathcal{O}(n^3)$ due to the matrix inversion. Finding probabilities for transitioning to a next state is $\mathcal{O}(n^2)$ because we already have matrix-inverse as a part of the state. Transitioning to the next state also takes $\mathcal{O}(n^2)$ because we apply Sherman-Morrison formula to update the inverse-transpose for the selected edge $j$.

\lstset{
    escapeinside={(*}{*)},          
}

\begin{algorithm}
\caption{Dependency tree model as a sequence generation model.}
\label{alg:state:transitions}
\begin{lstlisting}[language=Python]
def initial_state(W):
  i = 1
  L = L_hat(W)
  B = jnp.linalg.inv(L).T
  return i, W, L, B

def transition_probs(state)
  i, W, L, B = state
  return _for_a_given_B(W, B)[:,i]

def transit_state(state, j):
  Wp, Bp = ... (* apply Eq (\ref{eq:sherman:morrison:one:hot})-(\ref{eq:sherman:morrison:update:B}) *)
  return i+1, Wp, L_hat(Wp), Bp
\end{lstlisting}
\end{algorithm}

\section{Additional SWOR Results}
\label{sec:appendix:swor:results}

\begin{figure*}[h]
  \centering
  \subcaptionbox{30 words}{
    \includegraphics[width=0.31\textwidth]{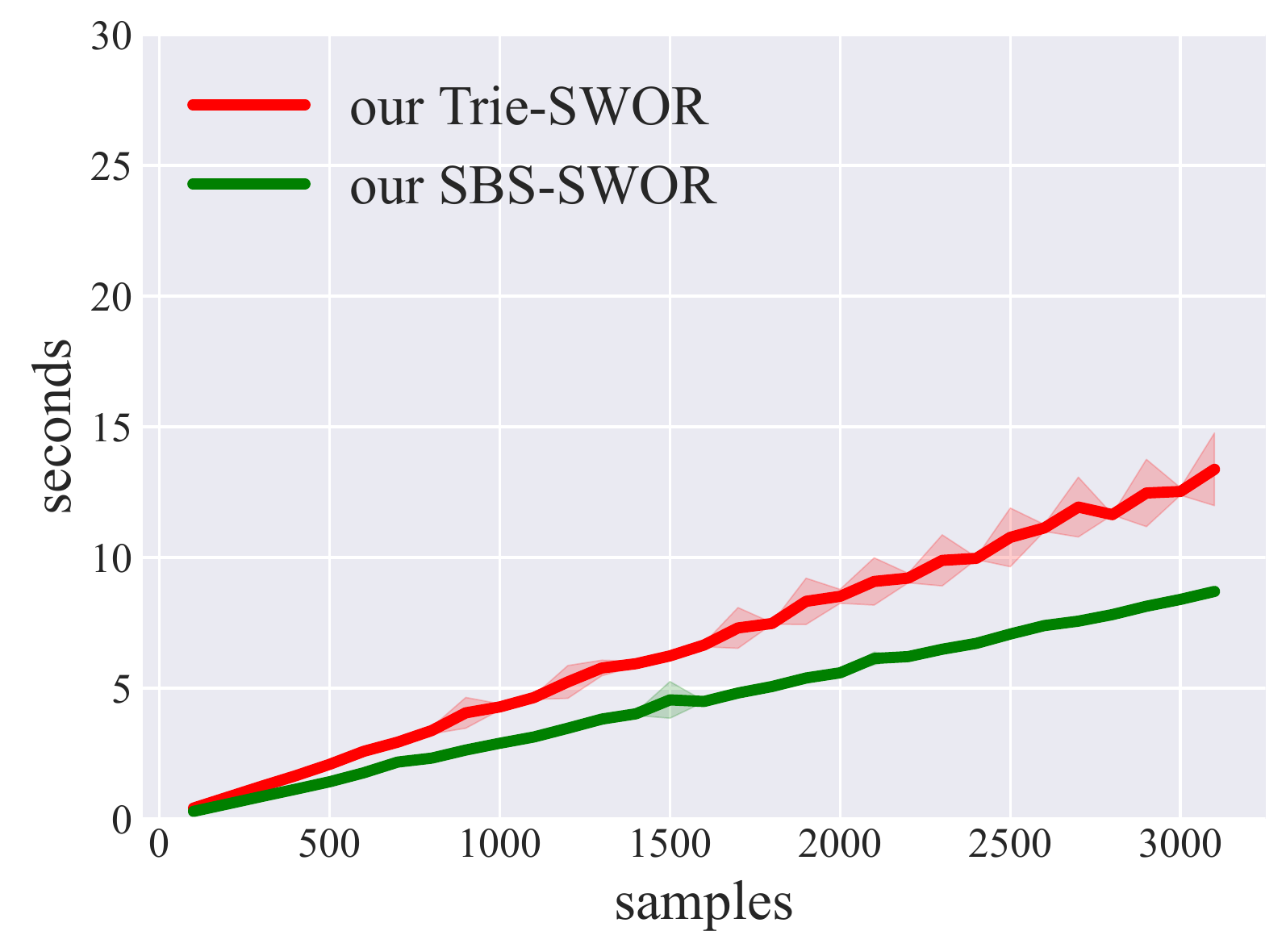}
  }\subcaptionbox{45 words}{
    \includegraphics[width=0.31\textwidth]{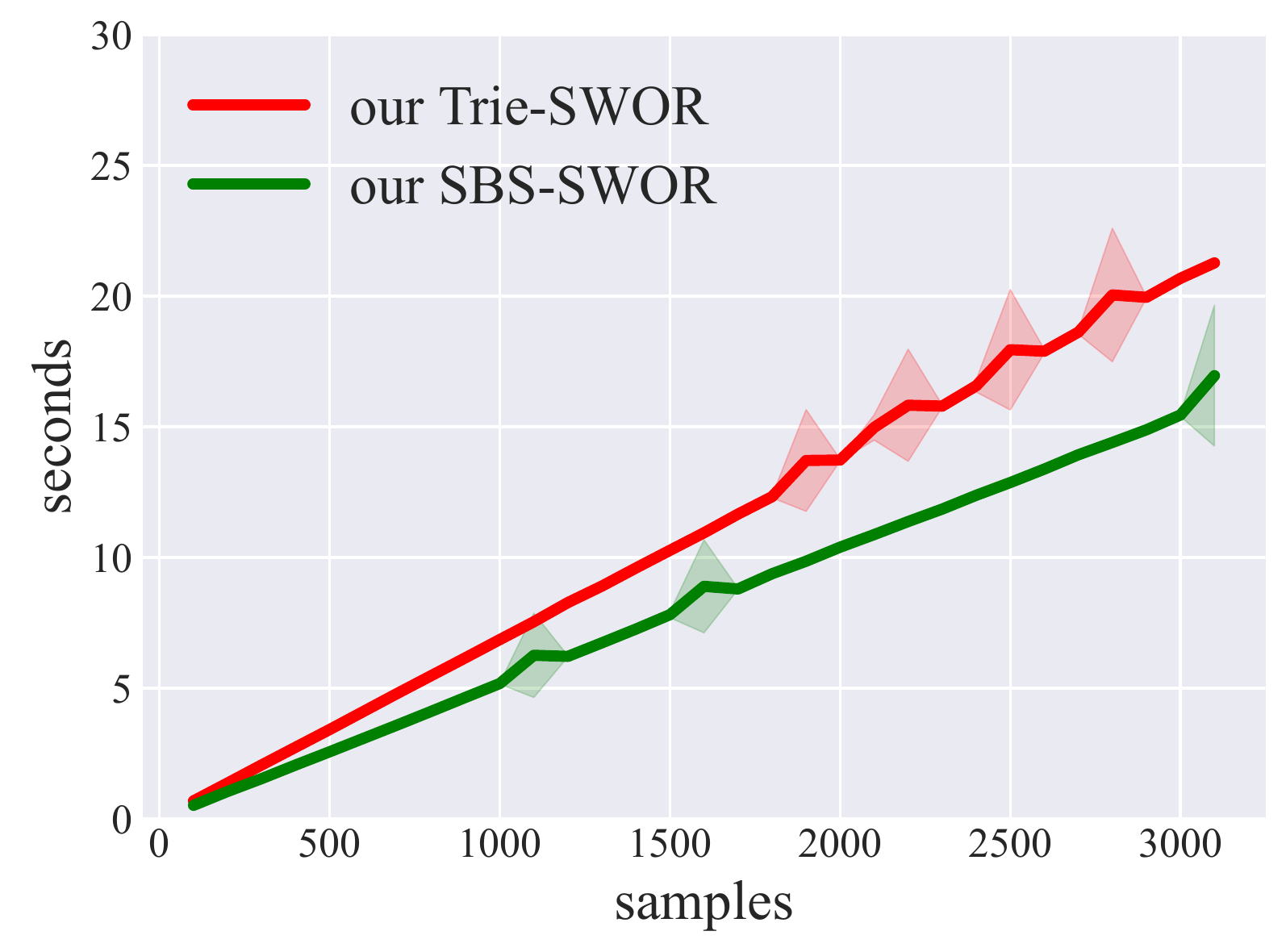}
  }\subcaptionbox{60 words}{
    \includegraphics[width=0.31\textwidth]{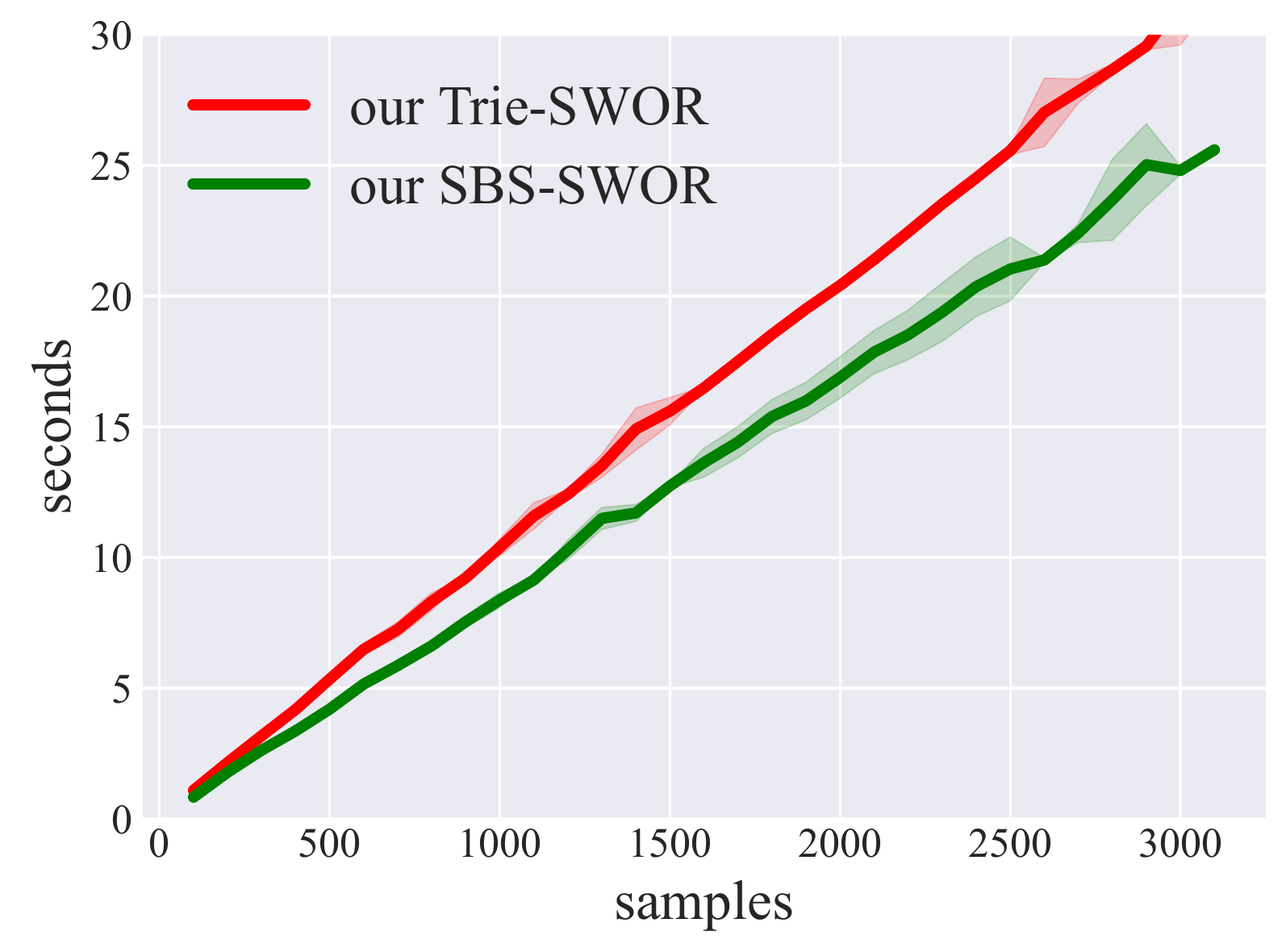}
  }
  \caption{Comparison of NumPy implementation of Trie-SWOR and SBS-SWOR algorithms over different sentence lengths.}
  \label{fig:swor:trie:sbs:many:lengths}
\end{figure*}

In the main text we have shown the results against the only previous SWOR sampler for dependency trees by \citet{zmigrod:sampling} for sentences with $14$ words. However, we could not run their algorithm for longer sentences so here we show only the results of our SWOR algorithms for different sentence length. In Figure~\ref{fig:swor:trie:sbs:many:lengths} are results of running the NumPy implementation of Trie-SWOR and SBS-SWOR and SBS-SWOR consistently outperforms Trie-SWOR due to the lower constant factors.

\begin{figure*}[h]
  \centering
  \subcaptionbox{30 words}{
    \includegraphics[width=0.31\textwidth]{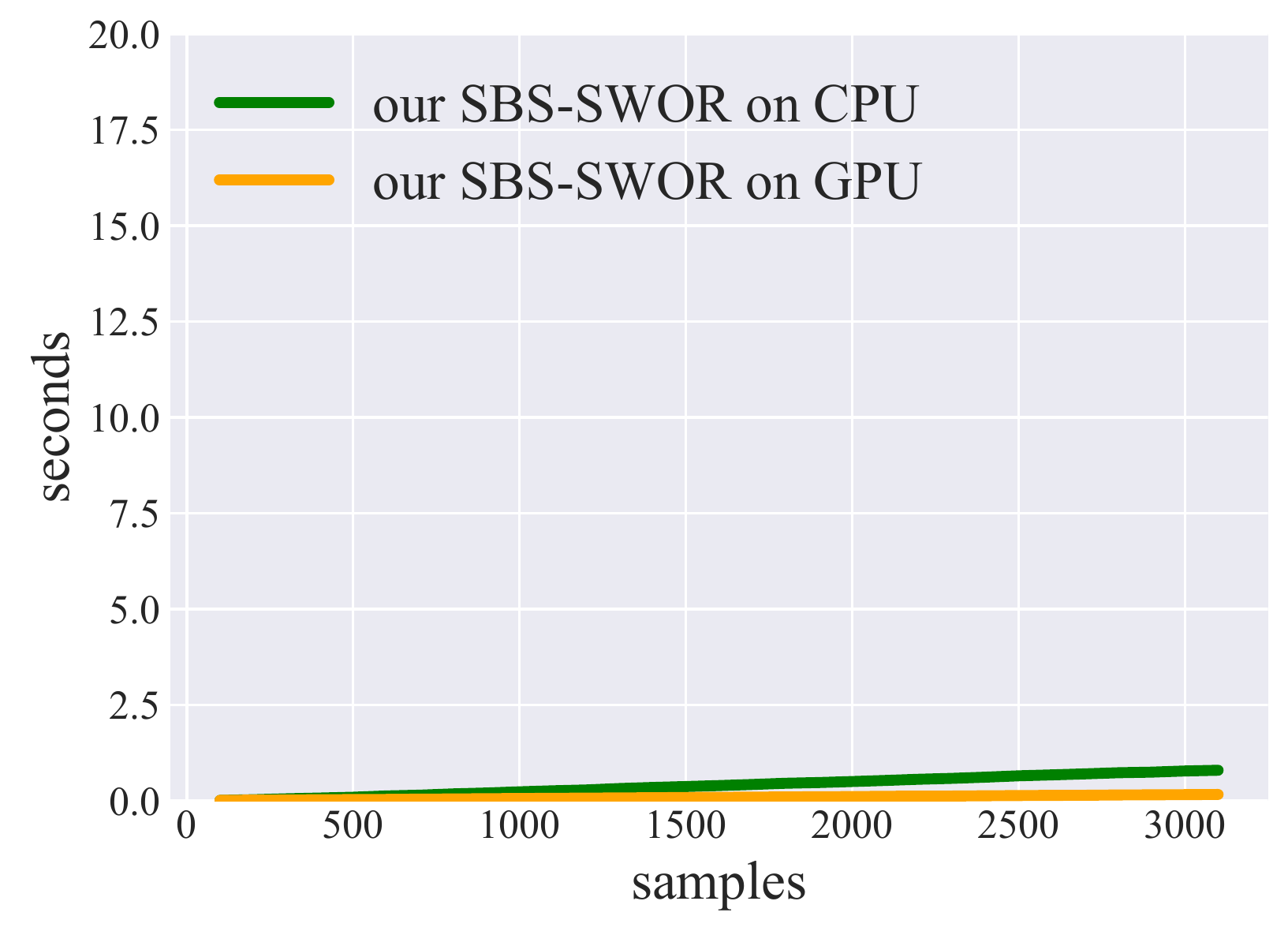}
  }\subcaptionbox{45 words}{
    \includegraphics[width=0.31\textwidth]{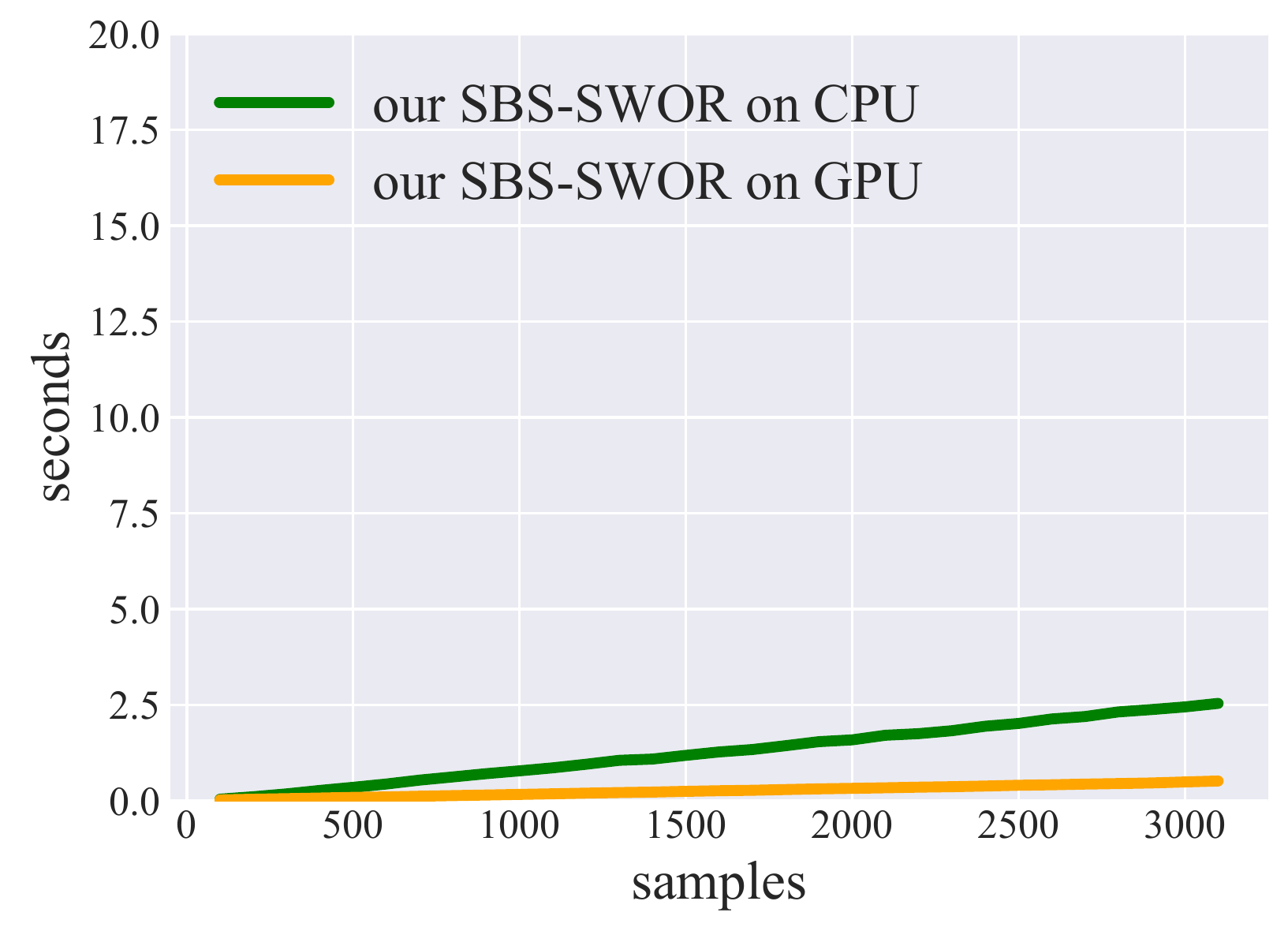}
  }\subcaptionbox{60 words}{
    \includegraphics[width=0.31\textwidth]{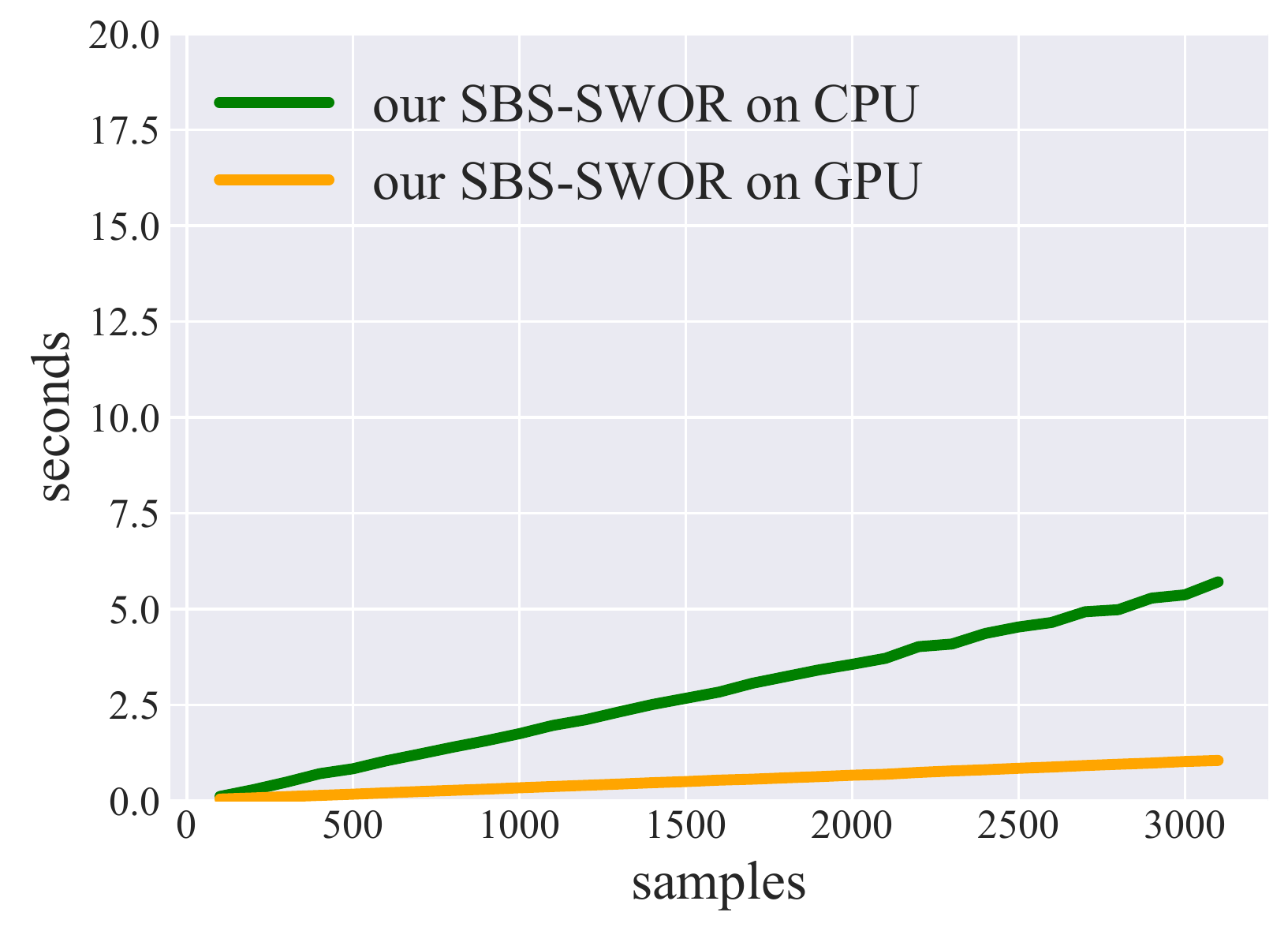}
  }
  \subcaptionbox{75 words}{
    \includegraphics[width=0.31\textwidth]{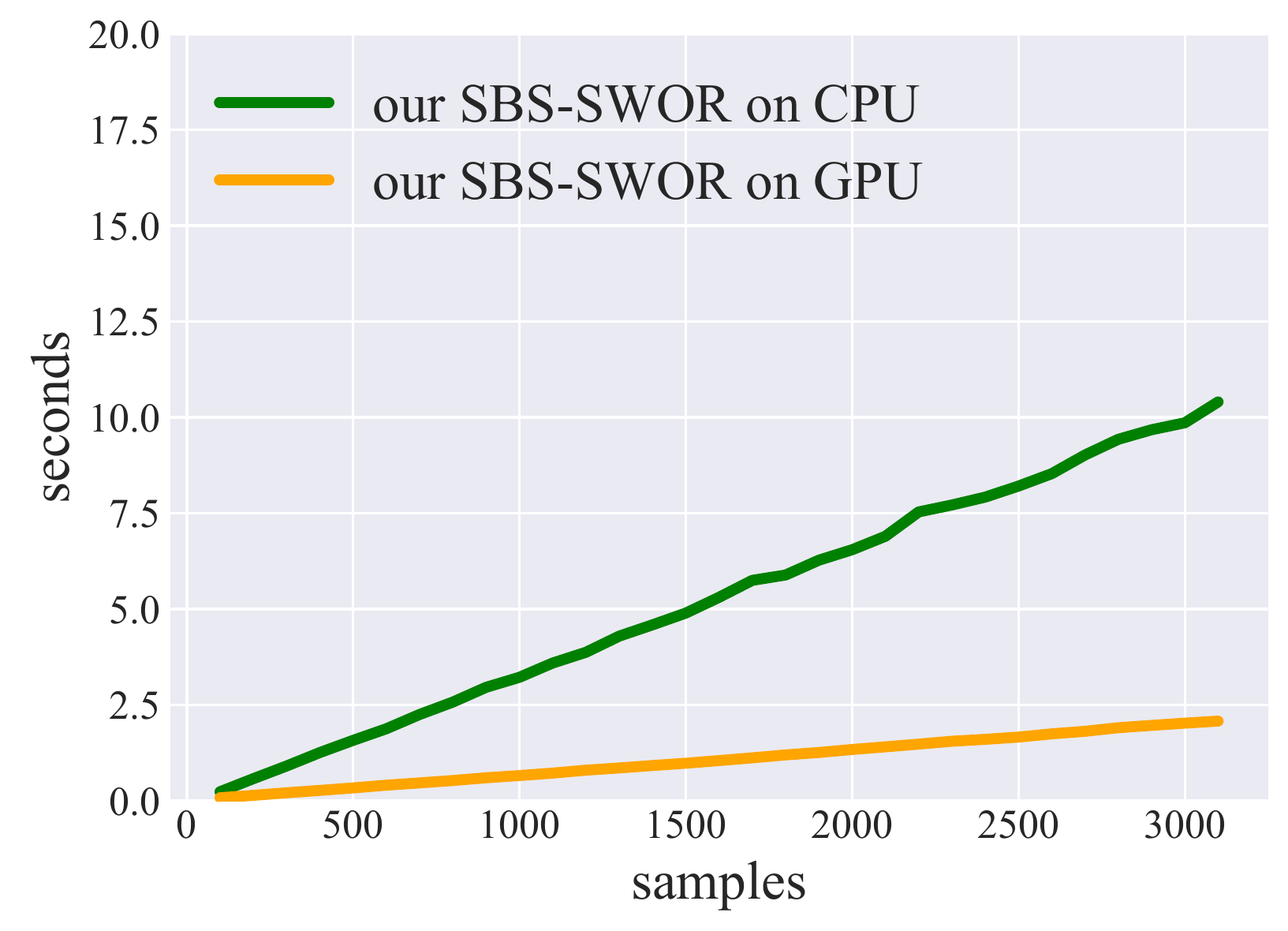}
  }\subcaptionbox{90 words}{
    \includegraphics[width=0.31\textwidth]{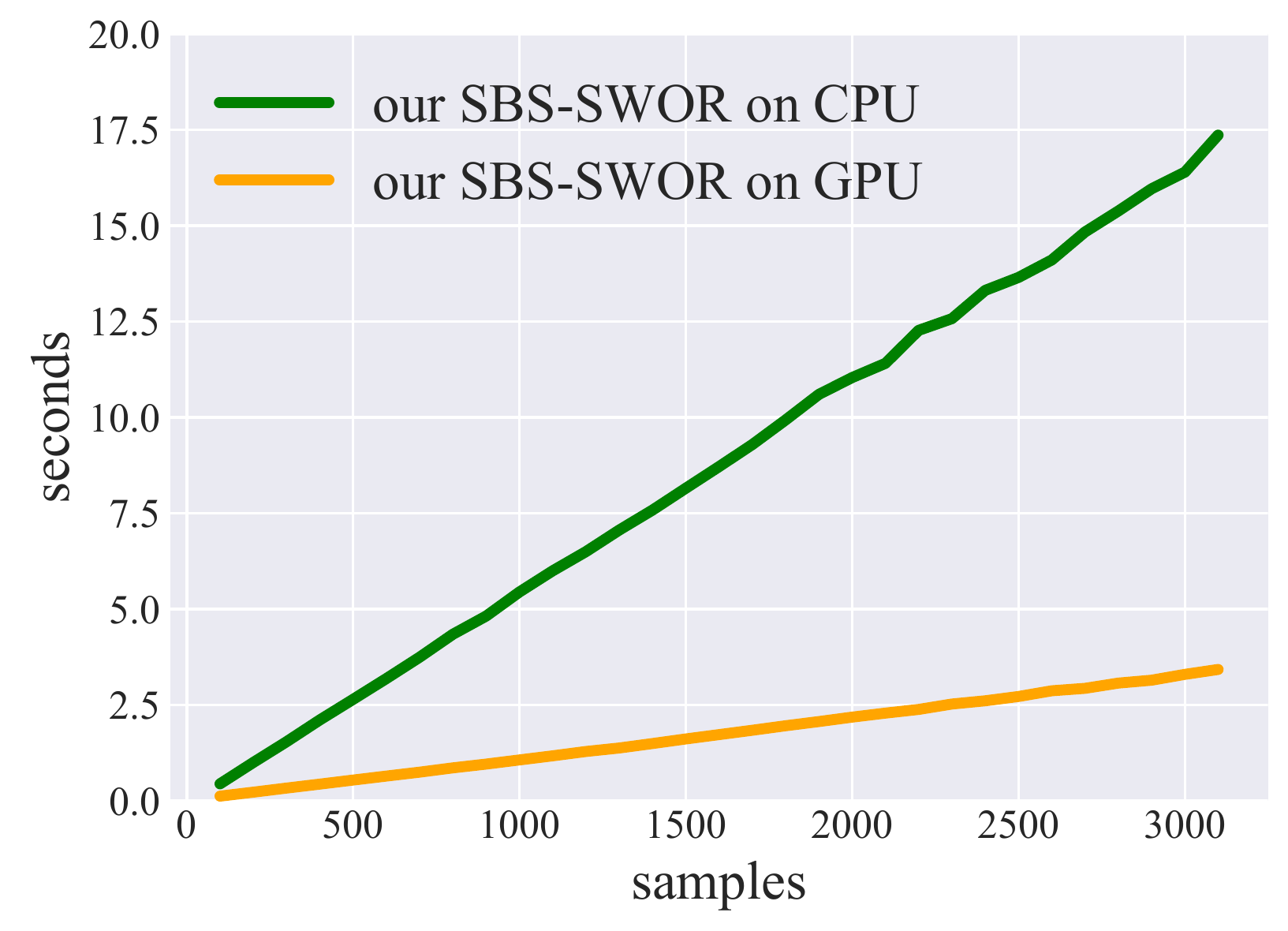}
  }\subcaptionbox{100 words}{
    \includegraphics[width=0.31\textwidth]{plots/plot_swor_jax_100.pdf}
  }
  \caption{Comparison of JAX implementation of SBS-SWOR on CPU vs GPU for different sentence lengths.}
  \label{fig:swor:sbs:GPU:many:lengths}
\end{figure*}

The real benefit of SBS-SWOR comes with parallel hardware. To show that we needed to reimplement the algorithm in some other toolkit that supports GPU execution. We used JAX as it has interface similar to NumPy. It is difficult to implement Trie-SWOR with JAX because JAX needs to know all shapes ahead of time, which is not possible to do with the dynamically growing data structure such as trie. We have implemented only SBS-SWOR and applied just-in-time compilation to it. The algorithm becomes substantially faster even on CPU, most likely because the compilation removes the Python overhead. Even such an optimized version that runs fast on CPU gets much faster if executed on GPU as visible from Figure~\ref{fig:swor:sbs:GPU:many:lengths}.
All experiments were ran on Intel Xeon$^\text{\textregistered}$ W-2135 CPU and NVIDIA$^\text{\textregistered}$ Quadro P1000 GPU.

\begin{figure*}[h!]
    \centering
    \includegraphics[width=1\textwidth]{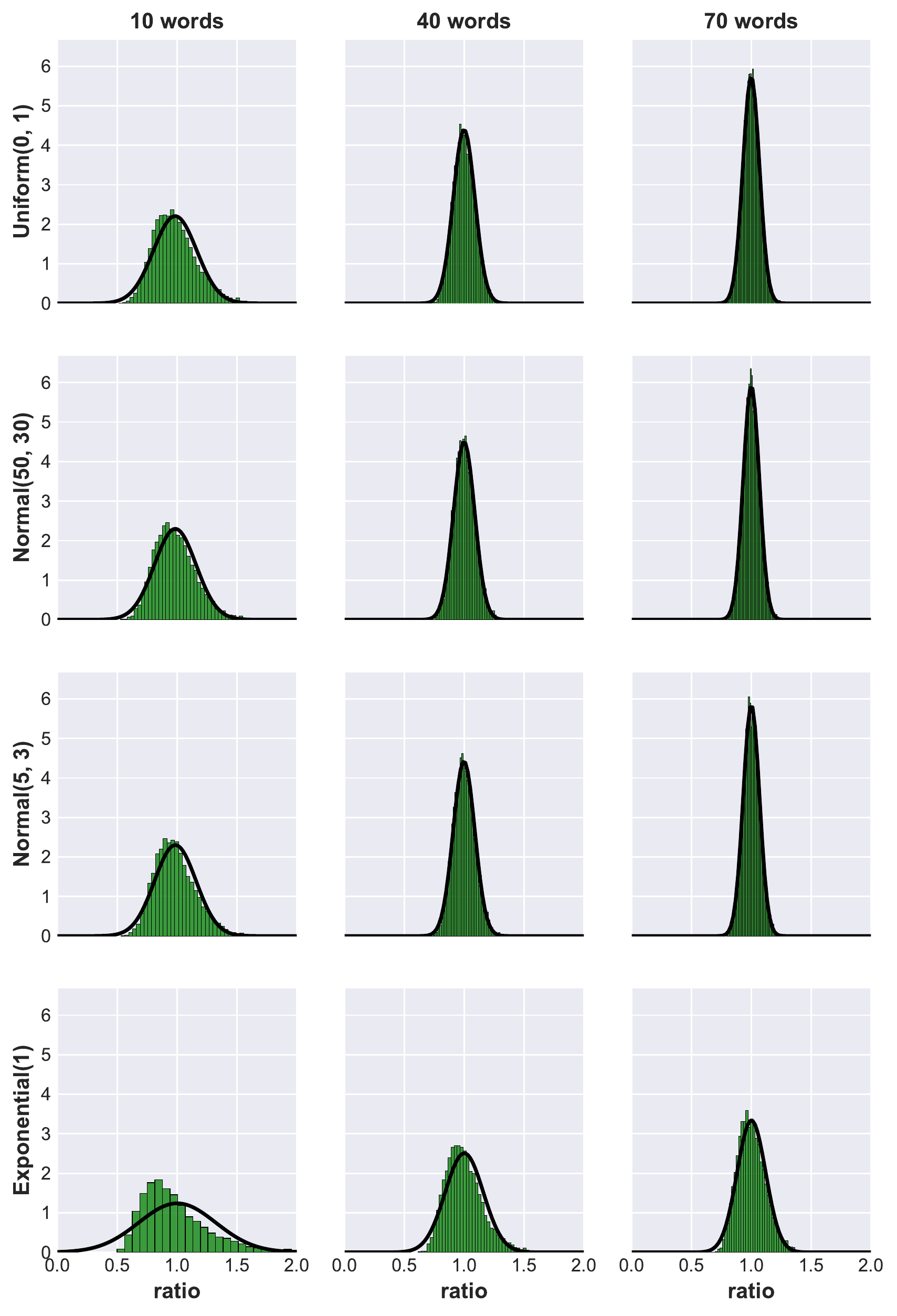}
    \caption{Distribution of ratio $\frac{w_{avg}^T}{w_{avg}^D}$ over graphs whose weights are sampled from different distributions.}
    \label{fig:average:ratios}
\end{figure*}

\section{Ratio of average tree weights}
\label{sec:appendix:assumption}

In Section~\switchCameraReady{sec:expected:runs:wilson:reject}{4.3.1} we have shown that with the first-order approximation of the expectation of ratio of dependency and spanning tree average weights, the expected number of trials of \textsc{WilsonReject} smaller than $e$. Here we give empirical support for it based on large number of simulations with different distributions and graph sizes.

To verify the approximation we have sampled $10,000$ weighted graphs for different combinations of weight distributions and sentence lengths and estimated ratio $\frac{w_{avg}^T}{w_{avg}^D}$. The results are shown in Figure~\ref{fig:average:ratios}. Since all the weights need to be non-negative we used the truncated version of Normal distribution, exponential distribution and uniform distribution. As can be seen from the results no matter what distribution we sample weights from and no matter what the sentence length is, the average value of the ratio $\frac{w_{avg}^T}{w_{avg}^D}$ is $\approx1$ which justifies the approximation made in Section~\switchCameraReady{sec:expected:runs:wilson:reject}{4.3.1}.

If we look only at the plots for the Truncated Normal distribution they look almost identical independently of the particular parameters that are used as mean and variance. Exponential seems to have a different shape of the distribution of ratio for small graph sizes, but the mean value of the ratio is still $1$.

Interestingly, for almost all sampled graphs the ratios of averages are smaller than $2$. 
If we make a weaker assumption that $\expectation{\frac{w_{avg}^T}{w_{avg}^D}}<2$ the main statement of Section~\switchCameraReady{sec:expected:runs:wilson:reject}{4.3.1} would still hold:
  \textsc{WilsonReject} is on average as fast as \textsc{Wilson} up to the small multiplicative constant.

\section{Formal Proof that \textsc{WilsonRC} is Biased}
\label{sec:appendix:wilsonrc:biased:proof}

In Section~\ref{sec:zmigrod:biased} we have showed that \textsc{WilsonRC} algorithm is biased through an example. Here we will show that formally and answer the question under what conditions can \textsc{WilsonRC} be unbiased. As a refresher, \textsc{WilsonRC} works as follows:
\begin{enumerate}
    \item sample one edge $e$ emanating from ROOT using their weights $w_e$,
    \item remove all ROOT outgoing edges except $e$,
    \item run regular Wilson's algorithm.
\end{enumerate}

The probability of sampling any dependency tree $t$ is given by the product of probability of sampling the root edge $e$ of $t$ in step $1$ and sampling $t$ in step $3$ from all the trees that remain after filtering in step $2$. If all the edges emanating from ROOT are given by set $R$, we can express the probability of sampling $t$ with \textsc{WilsorRC} as:

\begin{align}
    \frac{\phi(e)}{\sum_{e' \in R} \phi(e')} \frac{\phi(t)}{\sum_{t' \in D_e} \phi(t')} \label{eq:biasbias}
\end{align}

The sampling algorithm is unbiased if only if the probability of sampling any tree $t$ is given by $p(t) = \frac{\phi(t)}{Z}$ where $Z$ is a partition function. For \textsc{WilsonRC} to be unbiased all parts of Equation~\ref{eq:biasbias} except for $\phi(t)$ need to add up to $Z$:

\begin{align}
    \frac{\phi(e)}{\sum_{e' \in R} \phi(e')} \frac{1}{\sum_{t' \in D_e} \phi(t')} = \frac{1}{Z} \nonumber \\
    \iff \frac{\phi(e)}{\sum_{t' \in D_e} \phi(t')} = \frac{\sum_{e' \in R} \phi(e')}{Z}
    \label{eq:biasbias:two}
\end{align}

Since Equation~\ref{eq:biasbias:two} should hold for \textbf{all} root edges, and since the righthand side of the equation is constant for a given weighted graph, it follows that all root edges have the same value for the lefthand side of Equation~\ref{eq:biasbias:two}. If we pick two arbitrary root edges $e_1$ and $e_2$ we get:

\begin{align}
    \frac{\phi(e_1)}{\sum_{t' \in D_{e_1}} \phi(t')} = 
    \frac{\phi(e_2)}{\sum_{t' \in D_{e_2}} \phi(t')}
    \nonumber \\
    \iff 
    \frac{\phi(e_1)}{\phi(e_2)} = 
    \frac{\sum_{t' \in D_{e_1}} \phi(t')}{\sum_{t' \in D_{e_2}} \phi(t')}
    \nonumber \\
    \iff \frac{\phi(e_1)}{\phi(e_2)} = 
    \frac{p(e_1)}{p(e_2)}
    \nonumber
\end{align}

In other words, \textsc{WilsonRC} is unbiased if and only if the weights on the root edges happen to be exactly proportional to their marginal value which is an extremely unlikely scenario to happen with randomly sampled weight graph. For any other setting of weights \textsc{WilsonRC} returns biased samples. \textsc{WilsonMarginal} solves this problem by sampling the root edge directly from the marginal probability of the root edges, instead of using edge weights directly.

\end{document}